\documentclass[letterpaper,11pt]{article}

\usepackage{amssymb}
\usepackage{amsmath}
\usepackage{amsthm}
\usepackage{graphicx}
\usepackage{algorithm,,algpseudocode}
\usepackage{float}
\usepackage{booktabs,siunitx}
\usepackage{multirow}
\usepackage{fullpage}
\usepackage{bm}
\usepackage{physics}
\usepackage[super]{nth}
\usepackage[utf8x]{inputenc}
\usepackage{color}
\usepackage{algpseudocode, algorithm, algorithmicx}
\usepackage[T1]{fontenc}
\usepackage{romannum}
\usepackage{lineno}
\usepackage{caption}
\usepackage{subcaption}
\usepackage{cleveref}
\usepackage{bbm}
\usepackage{comment}
\usepackage{multicol}

\newtheorem{assumption}{Assumption}
\newtheorem{thm}{Theorem}

\newtheorem{lem}[thm]{Lemma}

\newtheorem{cor}[thm]{Corollary}

\newtheorem{rem}{Remark}

\newcommand{\inner}[1]{\left\langle #1\right\rangle}

\newcommand{\R}{\mathbb{R}}

\algnewcommand{\Inputs}{%
	\State \textbf{Inputs:}
}
\algnewcommand{\Initialize}{%
	\State \textbf{Initialize:}
}
\algnewcommand{\Outputs}{%
	\State \textbf{Outputs:}
}
\algnewcommand{\ForLoop}{%
	\textbf{For $j=1,2,...,m$ do}
}

\algnewcommand{\ForOuterLoop}{%
	\textbf{For $k=0,..., T-1$ do}
}

\algnewcommand{\ForOuterLoopOutput}{%
	\textbf{For $k=0,..., T'-1$ do}
}

\algnewcommand{\ForLoopKZPT}{%
	\textbf{For $j=0,2,..., p \lfloor {m \over p}\rfloor $ do}
}

\algnewcommand{\ForLoopmod}{%
	\textbf{For $L=0,1,...,L-1$ do}
}
\algnewcommand{\ForEnd}{%
	\textbf{End For}
}
\algnewcommand{\Iterate}{%
	\State \textbf{Iterate:}
}

\allowdisplaybreaks

\newcommand\amsclass[1]%
  {\hspace{9mm}
   \textbf{AMS subject classifications. }#1
}

\usepackage[T1]{fontenc}
\usepackage{romannum}
\usepackage{lineno}

\title{Robust Fourier Neural Networks}
\author{Halyun Jeong\thanks{clatar1@gmail.com, hjeong2@albany.edu} \hspace{1mm} and Jihun Han\thanks{jihun@math.dartmouth.edu}}
\date{}

\begin{document}
\pagenumbering{arabic}

\maketitle
\begin{abstract}
Fourier embedding has shown great promise in removing spectral bias during neural network training. However, it can still suffer from high generalization errors, especially when the labels or measurements are noisy. We demonstrate that introducing a simple diagonal layer after the Fourier embedding layer makes the network more robust to measurement noise, effectively prompting it to learn sparse Fourier features. We provide theoretical justifications for this Fourier feature learning, leveraging recent developments in diagonal networks and implicit regularization in neural networks. Under certain conditions, our proposed approach can also learn functions that are noisy mixtures of nonlinear functions of Fourier features. Numerical experiments validate the effectiveness of our proposed architecture, supporting our theory.
\end{abstract}



\section{Introduction}

We study the problem of learning a sparse sum of nonlinear functions of sinusoidals from noisy samples. More precisely, we consider the class of function $f$ of the following form
\begin{align*}
f(\theta) 
&= g_1(\cos (\pi a_1 \theta))  + g_2(\cos (\pi a_3 \theta)) + \dots + g_s(\cos (\pi a_{2s+1} \theta))\\
& \qquad + h_1(\sin (\pi a_2 \theta)) + h_2(\sin (\pi a_4 \theta))+\dots + h_{s'}(\sin (\pi a_{2 s'} \theta)),
\end{align*}
where $g_1$, \dots $g_s$ and $h_1$, \dots $h_{s'}$ are unknown  and (possibly) nonlinear link functions.
Here, $a_i \in [-m, m] \cap \mathbb{Z}$ are Fourier modes that govern the dominant frequencies of the function $f$. However, it does not restrict the function $f$ to be bandlimited to $[-m \pi, m \pi]$ due to the possible nonlinearity in the link functions $g_i$ and $h_i$.
Such functions include $f(\theta) = \cos(5\pi \theta) + \cos(21\pi \theta) + \sin(51\pi \theta)$ or more generally mixture of nonlinear functions of sinusodials such as $f(\theta) = (0.5\cos(5\pi x))^3 + \tanh (10 \cos(29\pi x)) + \max (\sin(61\pi x), 0)$.

Functions of this type commonly appear in machine learning and scientific computing since many datasets are cyclic or periodic in nature. For instance, due to the "day of the week effect", financial and search engine data often exhibit such patterns \cite{yang2022fourier}.

Reconstructing such functions from their noisy samples, represented as $f(\theta_i) + z_i$ for $i= 1, \dots, L$ where $z_i$ are random noise, poses several challenges: (1) The function may contain high-frequency components, complicating robust recovery due to the presence of noise. (2) Compared to the range of the frequency band of the function $f$, the actual frequencies governing the function are often sparse \cite{foucart2013invitation,laska2006random}. Identifying these true frequencies is crucial to reduce the generalization error and can be further utilized to unmix the mixture function $f$ \cite{heylen2014review}. (3) Many existing methods to sparsify the underlying features are based on estimating and pruning the estimated features while running the methods \cite{jin2016training,chen2021only}. Such approaches typically require prior information on the sparsity levels of $f$ (the number of governing frequencies) and they may not be available. In scenarios where this information is not readily available, setting the pruning parameter inaccurately might lead to unstable convergence or a slow convergence for these methods \cite{zhao2020optimal}.

Recently, neural networks with Fourier feature embedding \cite{tancik2020fourier} have been introduced to address the issue (1), capturing the high frequency parts of a target function better. Instead of using conventional positional encoding (standard basis), the authors employ Fourier features, showcasing their effectiveness through numerical experiments and neural tangent kernel (NTK) analysis \cite{jacot2018neural}. However, our numerical experiments indicate that Fourier embedding alone often suffers from overfitting when the noise is present in the samples. This observation also aligns with the philosophy of NTK since it works in the lazy training regime excluding the possibility of feature learning \cite{karp2021local, shi2023provable}.

On the other hand, there have been many recent works on the diagonal linear neural networks \cite{woodworth2020kernel,nacson2022implicit,even2023s}. Although this is equivalent to the reparametrization of the linear regression problem, it has been observed that its training dynamics differ, providing an implicit regularization effect that facilitates sparse feature learning under mild conditions.

Inspired by both Fourier embedding and the implicit regularization found in diagonal linear networks, we propose a novel neural network architecture that incorporates Fourier embedding followed by diagonal layers into ReLU networks, aiming to achieve the best of both worlds.

\subsection{Our contributions}

\begin{itemize}
    \item We introduce a novel architecture that merges Fourier embedding with diagonal neural networks. Both theoretical analysis and empirical experiments validate the efficacy of this scheme in learning a mixture of nonlinear functions of sinusoidals from noisy samples.
    
    \item We analyze the recently proposed Fourier feature neural networks beyond the kernel regime. This provides a clearer understanding of the inductive bias, elucidating its low generalization error and robustness to noise. A growing body of research suggests that the behavior of (stochastic) gradients in the kernel regime does not fully capture the inductive bias of ReLU neural networks \cite{mousavi2022neural, shi2023provable,even2023s}. Consequently, the original analysis in \cite{tancik2020fourier} might not comprehensively account for the observed empirical performance.

    \item Under mild conditions, for two layer diagonal neural networks with Fourier embedding with the identity and ReLU activation functions, we show that our approach recovers essential Fourier features (modes) of the target function $f(\theta) 
= g_1(\cos (\pi a_1 \theta))  + g_2(\cos (\pi a_3 \theta)) + \dots + g_s(\cos (\pi a_{2s+1} \theta)) + h_1(\sin (\pi a_2 \theta)) + h_2(\sin (\pi a_4 \theta))+\dots + h_{s'}(\sin (\pi a_{2 s'} \theta))$, even if the link functions $g_i$ and $h_i$ are unknown.

    \item Our method inherently identifies the sparsity pattern or Fourier modes of the target function. As such, our method often does not require hyperparameter tuning or knowledge of sparsity level of the true signal. This could be quite useful in practice since in many cases, such a knowledge is either unavailable or requires intricate hyperparameter tuning. 
    
\end{itemize}

\subsection{Related work}

While our work aligns in spirit with \cite{mousavi2022neural}, their theoretical framework does not directly apply to our context. Our design incorporates an additional embedding layer, and the components of its output are not statistically independent as theirs, which is crucial in their analysis. Actually, all the components of Fourier feature encoding are determined by the single parameter $\theta$.

The work by \cite{parkinson2023linear} proposes adding a linear layer to ReLU networks, but their proposed approach does not involve Fourier embedding, nor does it address the rate of convergence in feature learning or the approximation of the target function over the number of training iterations. In contrast, we study diagonal neural neural networks with Fourier embedding. Furthermore, our numerical experiments indicate that adding a diagonal layer with Fourier embedding to ReLU networks improves the generalization performance and robustness. 

Shi et al. \cite{shi2023provable} provides theoretical guarantees showing gradient descent learns important features, which improves the generalization error for two-layer neural networks. However, this work is for classification tasks not for the regression problem. Moreover, there is no consideration of Fourier embedding layer or diagonal layers. 

Oko et al. \cite{oko2024learning} investigate the problem of learning sums of link functions in single-index models. However, there are several key distinctions between their work and ours: (1) Our study focuses on the interaction between the Fourier embedding layer and diagonal layer in learning periodic nonlinear mixtures. (2) Their model assumes a componentwise independent Gaussian random vector as input, whereas in our case, the input is effectively a random vector derived from a one-dimensional input variable $\theta$ through the Fourier embedding layer, resulting in strong dependencies among the components. (3) These differences lead to key differences in our analytical approach. Specifically, we leverage Chebyshev and Fourier expansions and their associated properties, whereas their analysis is based on the Hermite expansion.

\subsection{Notation}
Let $\|\cdot\|_2$ and $\|\cdot\|_\infty$ represent the $\ell_2$ norm and the $\ell_\infty$ norm  for a vector respectively. For any positive integer $m$, the notation $[m]$ refers to the set of integers ${1, 2, \dots, m}$. The transpose of a matrix $A$ is written as $A^\top$. For positive semidefinite matrices $A$ and $B$, the notation $A \preceq B$ indicates that $B - A$ is positive semidefinite. 
The componentwise product of two vectors $v$ and $w$ of the same dimension is denoted by $v \odot w$. For any positive numbers $a$ and $b$, the notations $a \lesssim b$ and $a \gtrsim b$ indicate that there exists a universal constant $C > 0$ such that $a \le Cb$ and $a \ge Cb$, respectively. The activation function $\sigma$ is applied componentwise if it is applied to a vector. 

\subsection{Organization}
In Section 2, we introduce a diagonal neural network with Fourier embedding to learn the sum of nonlinear periodic mixtures from noisy samples, and a projected stochastic gradient for training our proposed networks. We then show that running a projected SGD on the diagonal layer enables the learning of the important Fourier modes of the target function under certain conditions. Section 3 discusses the existence of the second-layer weights for the target function, which achieve a low approximation error by utilizing the learned Fourier modes.
In Section 4, based on our findings in the previous two sections, we show that running another projected SGD for the second layer yields a trained model of the neural network with low generalization error. In Section 5, we validate the effectiveness of our approaches by conducting extensive numerical experiments using both synthetic and real-world data to learn the sum of nonlinear periodic mixtures. Section 6 provides discussions and conclusions on our work.

\section{Learning sparse features by Fourier diagonal networks}

Consider the noisy samples $y^k = \{ f(\theta^k)+z^k \}_{k=1}^K$, where $\theta^k$ are independent uniform random variables on $[-1,1]$ and $z^k$ are i.i.d. mean-zero bounded random noise that are independent with $\theta^k$. We focus on the class of functions $f$ with link functions $g_1$, \dots $g_s$ and $h_1$, \dots $h_{s'}$, defined as follows:
$f(\theta) 
= g_1(\cos (\pi a_1 \theta))  + g_2(\cos (\pi a_3 \theta)) + \dots + g_s(\cos (\pi a_{2s+1} \theta)) + h_1(\sin (\pi a_2 \theta)) + h_2(\sin (\pi a_4 \theta))+\dots + h_{s'}(\sin (\pi a_{2 s'} \theta))$.
We assume that $h_i$ are odd functions, motivated by considerations from the following approximation theory perspective.

First, note that $g_1(\cos (\pi a_1 \theta))  + g_2(\cos (\pi a_3 \theta)) + \dots + g_s(\cos (\pi a_{2s+1} \theta))$ is even for any nonlinear functions $g_i$ as $\cos (\pi a_i \theta)$ is even in $\theta$. On the other hand, given any function $f$, it can be decomposed as $f = f_{\text{even}} + f_{\text{odd}}$.
The even part $f_{\text{even}}$ can be approximated by the cosine series based on Fourier analysis, and in particular, it can be represented as a sum of $g_i(\cos (\pi i \theta))$ for appropriate nonlinear functions $g_i$.

Next, consider the odd part $f_{\text{odd}}$. From the decomposition of $h_i(\sin (\pi a_{2i} \theta))$ as
$h_i(\sin (\pi a_{2i} \theta)) = h_{i,\text{even}} (\sin (\pi a_{2i} \theta)) + h_{i,\text{odd}} (\sin (\pi a_{2i} \theta))$, one can easily see that
$h_{i,\text{even}} (\sin (\pi a_{2i} \theta))$ and $f_{\text{odd}}$ are orthogonal:
\[
\mathbb{E}_{\theta} [f_{\text{odd}} (\theta) h_{i,\text{even}} (\sin (\pi a_{2i} \theta))] = \int_{-1}^1 f_{\text{odd}} (\theta) h_{i,\text{even}} (\sin (\pi a_{2i} \theta)) d \theta = 0,
\] where the last step follows from the fact that the product of an even and an odd function is odd, and the integral of an odd function over a symmetric interval is zero. Hence, the odd function part of $f$ can only be approximated by considering $h_1(\sin (\pi a_2 \theta)) + h_2(\sin (\pi a_4 \theta))+\dots + h_{s'}(\sin (\pi a_{2 s'} \theta))$, where $h_i$ are odd link functions. 

Let $\alpha_{2i+1} := \mathbb{E}_{\theta}[f(\theta) \cos (\pi a_{2i+1} \theta)]$ for all $1 \le i \le s$ and $\alpha_{2i} := \mathbb{E}_{\theta}[f(\theta) \sin (\pi a_{2i} \theta)]$ for all $1 \le i \le s'$.

\begin{assumption}
\label{assumption:recovery_condition}
Consider the class of functions $f$ with link functions $g_1$, \dots $g_s$ and odd link functions $h_1$, \dots $h_{s'}$, defined as:
$f(\theta) 
= g_1(\cos (\pi a_1 \theta))  + g_2(\cos (\pi a_3 \theta)) + \dots + g_s(\cos (\pi a_{2s+1} \theta)) + h_1(\sin (\pi a_2 \theta)) + h_2(\sin (\pi a_4 \theta))+\dots + h_{s'}(\sin (\pi a_{2 s'} \theta))$. We assume that the link functions $g_1$, \dots $g_s$ and $h_1$, \dots $h_{s'}$ are Lipschitz continuous on $[-1,1]$. We also assume that the indices of the nonzero Fourier coefficients of each $g_l( \cos ( a_{2l+1} \theta))$ and $h_k( \sin ( a_{2k} \theta))$ do not overlap. \footnote{In other words, let $S_l = \{l_1, \dots, l_k \}$ be the subset of $[-m, m]$ corresponding to the nonzero coefficients of the Fourier series expansion $g_l( \cos ( a_{2l+1} x))$ or $h_l( \sin ( a_{2l} x))$. Then $S_l \cap S_j = \empty$ for $l \neq j$.} 
\end{assumption}

Since each link function $g(\cos (\pi a_i \theta))$ is Lipschitz continuous on $[-1,1]$,  it can be uniquely represented by its Chebyshev expansion, expressed in terms of Chebyshev polynomials of the first kind \cite{trefethen2019approximation, rivlin2020chebyshev}:
\[
g(w) = \lambda_0 +  \lambda_1 T_1(w) + \lambda_2 T_2(w) + \dots +,
\]
so
\begin{align}
\nonumber
g(\cos (\pi a_{2i +1} \theta)) 
\nonumber
&= \lambda_0 +  \lambda_1 T_1(\cos (\pi a_{2i +1} \theta)) + \lambda_2 T_2(\cos (\pi a_{2i +1} \theta)) + \dots +\\
\label{eq:link_function_expansion}
&= \lambda_0 +  \lambda_1 \cos (\pi 1 \cdot a_{2i +1} \theta) + \lambda_2 \cos (\pi 2 \cdot a_{2i +1} \theta) + \dots +, 
\end{align}
and in particular, 
\begin{align*}
&\mathbb{E}_{\theta}[g(\cos (\pi a_{2i +1} \theta)) T_1(\cos (\pi a_{2i +1} \theta))] = \mathbb{E}_{\theta}[g(\cos (\pi a_{2i +1} \theta)) \cos (\pi a_{2i +1} \theta)] = \lambda_1,
\end{align*}
where $T_i$ is the $i$-th Chebyshev polynomial. In particular, $T_1(w) = w$. Here, the equality \eqref{eq:link_function_expansion} follows from the well known property of Chebyshev polynomials of the first kind, $T_n(cos \theta) = cos (n\theta)$ \cite{trefethen2019approximation,rivlin2020chebyshev}. 

For the odd link function $h$, its expansion in Chebyshev polynomials is given by:
\[
h(w) = \zeta_0 +  \zeta_1 T_1(w) + \zeta_2 T_2(w) + \zeta_3 T_3(w) \dots +.
\]

Since the Chebyshev polynomials $T_{2k}$ are even functions, the coefficients $\zeta_0, \zeta_2, \zeta_4, \dots$ are zero. Using another property of Chebyshev polynomials that states $T_{2n+1}(sin \theta) = -sin((2n+1)\theta)$, we have
\begin{align}
\nonumber
h(\sin (\pi a_{2i} \theta)) 
\nonumber
&= \zeta_1 T_1(\sin (\pi a_{2i} \theta)) + \zeta_3 T_3(\sin (\pi a_{2i} \theta)) + \zeta_5 T_5(\sin (\pi a_{2i} \theta)) + \dots +\\
\label{eq:link_function_expansion2}
&=  -\zeta_1 \sin (\pi 1 \cdot a_{2i} \theta) - \zeta_3 \sin (\pi 3 \cdot a_{2i} \theta) - \zeta_5 \sin (\pi 5 \cdot a_{2i} \theta) - \dots -, 
\end{align}
and in particular, 
\begin{align*}
&\mathbb{E}_{\theta}[h(\sin (\pi a_{2i} \theta)) T_1(\sin (\pi a_{2i} \theta))] = -\mathbb{E}_{\theta}[h(\sin (\pi a_{2i} \theta)) \sin (\pi a_{2i} \theta)] = \zeta_1.
\end{align*}

\begin{rem} [Polynomial link functions]
\label{rem:mode_recovery_condition2}
    The above argument implies that if the link function $g$ or $h$ is a degree-$p$ polynomial, then the number of nonzero Fourier coefficients of each $g( \cos ( a_{2l+1} x))$ and $h( \sin ( a_{2k} x))$ is at most $p$ with the highest possible frequency $p \cdot a_{2l+1}$ or $p \cdot a_{2l}$. 
\end{rem}

\subsection{Two-layer diagonal neural networks with Fourier embedding}

The embedding function $\phi$ is the symmetrized version of the Fourier embedding \cite{tancik2020fourier}, defined as:
\begin{equation}
\phi(\theta)=\left[ 
\begin{matrix}
-\sin(\pi m\theta), -\cos(\pi m\theta), \dots, -\cos(\pi \theta), 1, \cos(\pi \theta),\sin(\pi \theta), \dots, \cos(\pi m\theta),\sin(\pi m\theta)
\end{matrix}
\right],
\end{equation}  where $\theta \in [-1,1]$.
Note that $\phi(\theta) \in \R^{4m +1}$. For the notational convenience, we index the elements of $\phi(\theta)$ by $\{-2m, -2m+1, \dots, -1, 0, 1, 2, \dots, 2m-1, 2m \}$. 

We consider the following two-layer diagonal neural networks with Fourier embedding $\vb x = \phi(\theta)$:
\begin{align}
\label{Two_layer_NN_output}
\hat{f}(\vb x; w, c) = \sum_{i=-2m}^{2m} c_i \sigma(w_i \vb x_i) = c^\top \sigma(w \odot \vb x), 
\end{align}
where $w$ is the weight vector of the diagonal layer, $c_i$ are the output layer weights, $\sigma$ is the ReLU activation function. Note that $w \odot \vb x$ in \eqref{Two_layer_NN_output} is componentwise product with two vectors, different from conventional matrix-vector product, since we employ the diagonal layer after Fourier embedding. 

Let us denote the $\ell_2$ loss function as $\ell (\hat{y}, y) = {1 \over 2} (\hat{y} - y)^2$. We denote the population risk $R$ and the empirical risk $\hat{R}$ as below.
\[
R(w, c) = \mathbb{E}_{\vb x, y} [\ell (\hat{f}(\vb x; w, c), y)], 
\]
\[
\hat{R}(w, c) = {1 \over TB} \sum_{k=1}^{TB} \ell (\hat{f}(\vb x^k; w, c), y^k). 
\]

We also consider the $\ell_2$ regularized loss function:
\[
R^\lambda (w, c) = \mathbb{E}_{\vb x, y} [\ell (f(\vb x; w, c), y)] + {\lambda \over 2} \|w\|_2^2 , \quad \hat{R}^\lambda (w, c) = {1 \over TB} \sum_{i=1}^{TB} \ell (f(\vb x^{(i)}; w, c), y^{(i)}) + {\lambda \over 2}  \|w\|_2^2. 
\]

\begin{rem}
    Note that the representation $\hat{f}(\vb x; w, c) = \sum\limits_{i=-2m}^{2m} c_i \sigma(w_i \vb x_i)$ in \eqref{Two_layer_NN_output} is inherently not unique for homogeneous activation functions such as the identity, ReLU, Leaky ReLU etc., due to the possible rescaling of $c_i$ and $w_i$ for each $i$. However, the product of $c_i w_i$ is unique. 
\end{rem}

\subsection{Analysis of the diagonal layer training}

We begin with a brief description of our proposed method based on the layer-wise training outlined in Algorithm \ref{alg:layerwise-training}. The layer-wise training approach is commonly used in recent neural network literature such as in \cite{mousavi2022neural,oko2024learning,lee2024neural}.

First, the first-layer parameters \( w \) are trained using stochastic gradient descent (SGD). In each iteration, the stochastic gradient \( g_w \) is computed from a fresh batch with size $B$, and the weights are updated by stepping in the opposite direction of the gradient and projecting the result onto the box \([-Q_1, Q_1]^{4m+1}\). This projection keeps the first-layer parameter weights to control their $\ell_\infty$ norm. After \( T \) iterations, the algorithm trains the second-layer parameters \( c \). Stochastic gradients \( g_c \) are computed similarly by subsampling data points, and the weights are updated and projected onto the box \([-Q_2, Q_2]^{4m+1}\).

This two-phase approach efficiently optimizes the model, utilizing the structure of each layer and maintaining control over parameter magnitudes. 

	\begin{algorithm} [ht]
	\caption{SGD: Layer-wise training}
	\begin{algorithmic}
		\Inputs $c^0, w^0 \in \R^m$, $\{\vb x^k, y^k\}_{k=1}^{TB}, T, T'$
		
		\Initialize the weights	$w$ and $c = c_0$.

        \ForOuterLoop
	       \begin{align*}
            &\text{$g_w := {1 \over B} \sum_{j=kB+1}^{(k+1)B} \nabla_w \ell(\hat{f}(\vb x^j; w^k, c^0), y^j)) + \lambda w^k$} \quad \text{[Stochastic gradient of $\hat{R}^\lambda$]}\\
	        &u^{k} := w^k - \eta_k g_w \\
            &w^{k+1} := P_{Q_1}\left(u^{k} \right) \qquad \text{[Projection onto the box $[-Q_1,Q_1]^{4m+1}$}]
	       \end{align*}
           
		\indent \ForEnd

		\ForOuterLoopOutput
        \begin{align*}
            &\text{$g_c :=$ Stochastic gradient of $\hat{R}$ in $c$ by subsampling $B$ data points}\\
	        &v^k := c^k - \eta'_k g_c \\
            &c^{k+1} := P_{Q_2}\left(v^k \right) \qquad \text{[Projection onto the box $[-Q_2,Q_2]^{4m+1}$}]
	       \end{align*}
		\indent \ForEnd
		\Outputs $w^{T}, c^{T'}$
		
	\end{algorithmic}
	\label{alg:layerwise-training}
\end{algorithm}

To analyze Algorithm \ref{alg:layerwise-training}, we start with the gradient of the population risk $R$ with respect to the first layer weight $w$.
\begin{align}
\nonumber
& \nabla_{w_j} \mathbb{E}_{\vb x, y} [ \ell (\hat{f}(\vb x; w, c), y)] \\
\nonumber
&= \mathbb{E}_{\vb x, y} [\nabla_{w_j} \ell (\hat{f}(\vb x; w, c), y)]  \\
\nonumber
&= \mathbb{E}_{\vb x, y} [ (\hat{f}(\vb x; w, c) - y)  ( (c_j \sigma'(w_j x_j))  x_j) ]\\
\nonumber
&= \mathbb{E}_{\vb x, y} [ \hat{f}(\vb x; w, c) \cdot c_j \sigma'(w_j x_j))  x_j ] - \mathbb{E}_{\vb x, y} [ y(\vb x)   (c_j \sigma'(w_j x_j))  x_j]\\
\nonumber
&= c_j \mathbb{E}_{\vb x, y} \left[  \sum_{k=-2m}^{2m}  c_k \sigma(w_k x_k)   \cdot  \sigma'(w_j x_j))  x_j \right]  - \mathbb{E}_{\vb x, y} [ y(\vb x)   (c_j \sigma'(w_j x_j))  x_j]\\
\label{eq:population_gradient_first_layer}
&= c_j  \sum_{k=-2m}^{2m}  c_k \mathbb{E}_{\vb x, y} \left [  \sigma(w_k x_k) \sigma'(w_j x_j) x_j  \right] - \mathbb{E}_{\vb x, y} [ y(\vb x) \cdot c_j \sigma'(w_j x_j) x_j  ].
\end{align}

Using the above computation, the gradient of the regularized population risk $R^\lambda$ is given by
\small
\begin{align*}
& \nabla_{w_j} R^\lambda(w,c)\\
&= \sum_{k=-2m}^{2m}  c_j c_k \mathbb{E}_{\vb x, y} \left [  \sigma(w_k x_k) \sigma'(w_j x_j) x_j  \right] - \mathbb{E}_{\vb x, y} [ y(\vb x) \cdot c_j \sigma'(w_j x_j) x_j  ] + \lambda w_j\\
&= \sum_{k=-2m}^{2m}  c_j c_k |w_k|  \mathbb{E}_{\vb x, y} \left [  \sigma(\text{sign}(w_k) x_k) \sigma'(\text{sign}(w_j) x_j)  x_j  \right] - \mathbb{E}_{\vb x, y} [ y(\vb x) \cdot c_j \sigma'(\text{sign}(w_j) x_j) x_j  ] + \lambda w_j\\
&= c_j \sum_{k=-2m}^{2m}   c_k \text{sign}(w_j) |w_k|  \mathbb{E}_{\vb x, y} \left [  \sigma(\text{sign}(w_k) x_k) \sigma'(\text{sign}(w_j) x_j)  \text{sign}(w_j) x_j  \right] \\
&\qquad -  c_j  \mathbb{E}_{\vb x, y} [ y(\vb x) \cdot\sigma'(\text{sign}(w_j) x_j) x_j  ] + \lambda w_j\\
&= c_j \sum_{k=-2m}^{2m}  c_k \text{sign}(w_j) \text{sign}(w_k) w_k  \mathbb{E}_{\vb x, y} \left [  \sigma(\text{sign}(w_k) x_k) \sigma'(\text{sign}(w_j) x_j)  \text{sign}(w_j) x_j  \right] \\
&\qquad - c_j \mathbb{E}_{\vb x, y} [ y(\vb x) \cdot  \sigma'(\text{sign}(w_j) x_j) x_j  ] + \lambda w_j\\
&= c_j \sum_{k=-2m}^{2m}  c_k \text{sign}(w_j) \text{sign}(w_k) w_k  \mathbb{E}_{\vb x, y} \left [  \sigma(\text{sign}(w_k) x_k) \sigma(\text{sign}(w_j) x_j)  \right] \\
&\qquad - c_j \mathbb{E}_{\vb x, y} [ y(\vb x) \cdot  \sigma'(\text{sign}(w_j) x_j) x_j  ] + \lambda w_j\\
&= c_j \sum_{k=-2m}^{2m}  c_k \text{sign}(w_j) \text{sign}(w_k) w_k  \mathbb{E}_{\vb x, y} \left [  \sigma(\text{sign}(w_k) x_k) \sigma(\text{sign}(w_j) x_j)  \right] \\
&\qquad - c_j \text{sign}(w_j) \mathbb{E}_{\vb x, y} [ y(\vb x) \cdot  \sigma'(\text{sign}(w_j) x_j) \text{sign}(w_j) x_j  ] + \lambda w_j\\
&= c_j \text{sign}(w_j) \sum_{k=-2m}^{2m}  c_k |w_k|  \mathbb{E}_{\vb x, y} \left [  \sigma(\text{sign}(w_k) x_k) \sigma(\text{sign}(w_j) x_j)  \right] \\
&\qquad - c_j \text{sign}(w_j) \mathbb{E}_{\vb x, y} [ y(\vb x) \cdot  \sigma(\text{sign}(w_j) x_j) ] + \lambda w_j.
\end{align*}
\normalsize

Thus, we obtain 
\begin{align}
\label{eq:gradient_first_layer}
\nabla_{w} R^\lambda(w,c) &=   c \odot \text{sign}(w) \odot h(w) - c \odot  \text{sign}(w) \odot \text{vec}_j \left(\mathbb{E}_{\vb x, y} [ y(\vb x) \cdot  \sigma(\text{sign}(w_j) x_j)  ] \right) + \lambda w,
\end{align}
where $h(w) $ is a vector whose $j$-entry is defined as 
\begin{align}
\nonumber
h(w)_{j}
& :=   \sum_{k=-2m}^{2m}  c_k |w_k|  \mathbb{E}_{\vb x, y} \left [  \sigma(\text{sign}(w_k) x_k) \sigma(\text{sign}(w_j) x_j)  \right] \\  
\nonumber
&\le   \sum_{k=-2m}^{2m} c_k |w_k| \\
\label{eq:bound_h}
&\le \|c\|_2 \|w\|_2,
\end{align}
where the first inequality is from $|x_i| \le 1$ since $x_i$ are sinusoids and the second inequality is from the Cauchy-Schwartz inequality. 

From the first loop in Algorithm \ref{alg:layerwise-training} for training the first layer, we have
\small
\begin{align*}
&u^t\\ 
&= w^t - \eta_t g_w\\
&= w^t - \eta_t \nabla_{w} R^\lambda(w,c) - \eta_t \left(g_w - \nabla_{w} R^\lambda(w,c) \right)\\
&= w^t - \eta_t \nabla_{w} R^\lambda(w,c) - \eta_t \Gamma_t\\
&= w^t - \eta_t \left( c \odot \text{sign}(w^t) \odot h(w^t) - c \odot  \text{sign}(w^t) \odot \text{vec}_j \left(\mathbb{E}_{\vb x, y} [ y(\vb x) \cdot  \sigma(\text{sign}(w^t_j) x_j)  ] \right) + \lambda w^t \right)   - \eta_t \Gamma_t\\
&= \left( I- \eta_t \lambda I  \right)w^t - \eta_t c \odot \text{sign}(w^t) \odot h(w^t) + \eta_t   c \odot  \text{sign}(w^t) \odot \text{vec}_j \left(\mathbb{E}_{\vb x, y} [ y(\vb x) \cdot  \sigma(\text{sign}(w^t_j) x_j)  ] \right)   - \eta_t \Gamma_t\\
&= \left( I- \eta_t \lambda I  \right)w^t - \eta_t c \odot \text{sign}(w^t) \odot h(w^t) + \eta_t   c \odot  \text{sign}(w^t) \odot \text{vec}_j \left(\mathbb{E}_{\vb x, z} [ (f(\vb x) + z) \sigma(\text{sign}(w^t_j) x_j)  ] \right)   - \eta_t \Gamma_t\\
&= \left( I- \eta_t \lambda I  \right)w^t - \eta_t c \odot \text{sign}(w^t) \odot h(w^t) + \eta_t   c \odot  \text{sign}(w^t) \odot \text{vec}_j \left(\sum_{l=0}^{2m} \widetilde{\alpha}_l \mathbb{E}_{\vb x   } \left[ x_l\sigma \left(  \text{sign}(w^t_j) x_j \right)  \right] \right)   - \eta_t \Gamma_t,
\end{align*}
\normalsize
where $\text{vec}_j$ is understood as a vector with its components indexed by $j$. $\Gamma_t$ represents the stochastic gradient noise, which is the difference between the stochastic gradient $g_w$ and the gradient of $\nabla_{w} R^\lambda(w,c)$ in $w$. 

To proceed further, we need the following technical assumption for the Fourier mode recovery. 
\begin{assumption}
\label{assumption:recovery_condition2}
Suppose the target function $f$ has a sine-cosine series expansion $\sum_{l=0}^{2m} \widetilde{\alpha}_l x_l$ with possible nonzero coefficients indexed by either the form $l = 4k+1$ or $l = 4k+2$. \footnote{Although this condition seems restrictive, most of related works for recovery using two-layer NN requires such conditions. See for example Assumption 3 in \cite{lee2024neural}.}
\end{assumption}
This leads to the following lemma. 
\begin{lem}
Suppose Assumption \ref{assumption:recovery_condition2} holds. Then, we have 
\begin{align*}
&\sum_{l=0}^{2m} \widetilde{\alpha}_l \mathbb{E}_{\vb x   } \left[ x_l\sigma \left(  \text{sign}(w^t_j) x_j \right)  \right] = \text{sign}(w^t_j) \cdot {\widetilde{\alpha}_r \over 2}
\end{align*}
or
\begin{align*}
\text{vec}_j \left(\sum_{l=0}^{2m} \widetilde{\alpha}_l \mathbb{E}_{\vb x   } \left[ x_l\sigma \left(  \text{sign}(w^t_j) x_j \right)  \right] \right) = \text{sign}(w^t) \odot  { \alpha \over 2}.
\end{align*}
\end{lem}

\begin{proof}
The Chebyshev polynomial expansion of the ReLU activation function $\sigma$ (See Appendix A in \cite{fanaskov2022spectral}),
$\sigma(\cos(\pi j \theta)) = \sum_{i=0}^\infty p_i \cos(\pi ij \theta)$, where
$$
p_i=
\begin{cases}
{1 \over \pi}, \quad \text{for $i = 0$}, \\
{1 \over 2}, \quad \text{for $i = 1$},\\
{2 \over \pi} {\cos(i \pi/2) \over 1 - i^2} \quad \text{for $i \ge 2$}.
\end{cases}
$$
We have the following observations: when $i \ge 3$ is an odd number, then $p_i = 0$.

Suppose $\text{sign}(w^t_j) = 1$ and $j=2r-1$ where $r$ is a positive integer. 
\begin{align*}
&\sum_{l=0}^{2m} \widetilde{\alpha}_l \mathbb{E}_{\vb x   } \left[ x_l\sigma \left(  \text{sign}(w^t_j) x_j \right)  \right] \\
&= \sum_{l=0}^{2m} \widetilde{\alpha}_l \mathbb{E}_{\vb x   } \left[ x_l\sigma \left( x_j \right)  \right]\\
&= \sum_{l=1}^{m} \widetilde{\alpha}_{2l-1} \mathbb{E}_{\vb x   } \left[  \cos(\pi l \theta)  \sum_{i=0}^\infty p_i \cos(\pi ir \theta) \right] + \sum_{l=1}^{m} \widetilde{\alpha}_{2l} \mathbb{E}_{\vb x   } \left[ \sin(\pi l \theta) \sum_{i=0}^\infty p_i \cos(\pi ir \theta) \right]  \\
&= \sum_{l=1}^{m} \widetilde{\alpha}_{2l-1} \mathbb{E}_{\vb x   } \left[  \cos(\pi l \theta)  \sum_{i=0}^\infty p_i \cos(\pi ir \theta) \right]  \\
&= \sum_{l=1}^{m} \widetilde{\alpha}_{2l-1} \sum_{i=0}^\infty p_i \mathbb{E}_{\vb x   } \left[ \cos(\pi l \theta) \cos(\pi ir \theta) \right] .
\end{align*}
The second last inequality is from $\mathbb{E}_{\vb x   } \left[  \sin(\pi l \theta)   \cos(\pi ir \theta) \right] = 0$. 
Note that $\mathbb{E}_{\vb x   } \left[ \cos(\pi l \theta) \cos(\pi ir \theta) \right] = 0$ unless $l = ir$ for some integer $i$. Since $\widetilde{\alpha}_l$ with only indices of the form $l = 4k+1$ or $l = 4k+2$ can be nonzero, $\widetilde{\alpha}_{2l-1}$ can be nonzero only if $l$ is odd. Thus, the only possible nonzero terms $\mathbb{E}_{\vb x   } \left[ \cos(\pi l \theta) \cos(\pi ir \theta) \right]$ are those with $i$ being odd. Since $p_i = 0$ for all odd $i \ge 3$, this makes the inner sum $\sum\limits_{i=0}^\infty p_i \mathbb{E}_{\vb x   } \left[ \cos(\pi l \theta) \cos(\pi ir \theta) \right] = 0$ unless $l = r$ or $i = 1$.
Thus, we have
\begin{align*}
&\sum_{l=0}^{2m} \widetilde{\alpha}_l \mathbb{E}_{\vb x   } \left[ x_l\sigma \left(  \text{sign}(w^t_j) x_j \right)  \right] = \widetilde{\alpha}_{2r-1} p_1 \mathbb{E}_{\vb x   } \left[ \cos(\pi r \theta)^2 \right] =   \widetilde{\alpha}_{2r-1}  p_1 = \text{sign}(w^t_j) \cdot {\widetilde{\alpha}_j \over 2}.
\end{align*}
When  $\text{sign}(w^t_j) = -1$ and $j=2r-1$, 
\begin{align*}
\sigma(\text{sign}(w^t_j)  \cos(\pi r \theta)) &= \sigma(-\cos(\pi r \theta))\\
&= \sigma(\cos(\pi (r \theta + 1))) \\
&= \sum_{i=0}^\infty p_i \cos(\pi i(r \theta + 1))\\
&= \sum_{i=0}^\infty p_i (-1)^i \cos(\pi ir \theta)).
\end{align*}
Again, $\mathbb{E}_{\vb x   } \left[ \cos(\pi l \theta) \cos(\pi ir \theta) \right] = 0$ unless $l = ir$, making the inner sum $\sum\limits_{i=0}^\infty p_i \mathbb{E}_{\vb x   } \left[ \cos(\pi l \theta) \cos(\pi ir \theta) \right] = 0$ unless $l = r$ or $i = 1$. Thus, 
\begin{align*}
&\sum_{l=0}^{2m} \widetilde{\alpha}_l \mathbb{E}_{\vb x   } \left[ x_l\sigma \left(  \text{sign}(w^t_j) x_j \right)  \right] = - \widetilde{\alpha}_{2r-1} p_1 \mathbb{E}_{\vb x   } \left[ \cos(\pi r \theta)^2 \right] =  - \widetilde{\alpha}_{2r-1}  p_1 = \text{sign}(w^t_j) \cdot  {\widetilde{\alpha}_j \over 2}.
\end{align*}

The case for $\text{sign}(w^t_j) = 1$ and $j=2r$ where $r$ is a positive integer can be argued similarly. Note that 
\begin{align*}
\sigma(\text{sign}(w^t_j)  \sin(\pi r \theta)) &= \sigma(\sin (\pi r \theta))\\
&= \sigma(\cos(\pi (r \theta + 3/2))) \\
&= \sum_{i=0}^\infty p_i \cos(\pi i(r \theta + 3/2)).
\end{align*}
Thus, by a similar reasoning used in the above argument, we have
\small
\begin{align*}
&\sum_{l=0}^{2m} \widetilde{\alpha}_l \mathbb{E}_{\vb x   } \left[ x_l\sigma \left(  \text{sign}(w^t_j) x_j \right)  \right] \\
&= \sum_{l=1}^{m} \widetilde{\alpha}_{2l-1} \mathbb{E}_{\vb x   } \left[  \cos(\pi l \theta)  \sum_{i=0}^\infty p_i \cos(\pi i(r \theta + 3/2))\right] + \sum_{l=1}^{m} \widetilde{\alpha}_{2l} \mathbb{E}_{\vb x   } \left[ \sin(\pi l \theta) \sum_{i=0}^\infty p_i \cos(\pi i(r \theta + 3/2)) \right]  \\
&= \sum_{l=1}^{m} \widetilde{\alpha}_{2l} \mathbb{E}_{\vb x  } \left[ \sin(\pi l \theta) \sum_{i=0}^\infty p_i \cos(\pi i(r \theta + 3/2)) \right]  \\
&= \sum_{l=1}^{m} \widetilde{\alpha}_{2l} \sum_{i=0}^\infty p_i \mathbb{E}_{\vb x   } \left[ \sin(\pi l \theta)  \cos(\pi i(r \theta + 3/2))  \right] \\
&= \text{sign}(w^t_j) \cdot  {\widetilde{\alpha}_j \over 2}.
\end{align*}
\normalsize
The remaining case $\text{sign}(w^t_j) = -1$ and $j=2r$ can be handled similarly. 
For $j \in [-2m, -1]$, since $x_j = - x_{-j}$, 
$\sum_{l=0}^{2m} \widetilde{\alpha}_l \mathbb{E}_{\vb x   } \left[ x_l\sigma \left(  \text{sign}(w^t_j) x_j \right)  \right] = - \text{sign}(w^t_j) \cdot  {\widetilde{\alpha}_j \over 2}$.
\end{proof}

Hence, we have
\small
\begin{align}
&u^t \\
\nonumber
&= \left( I- \eta_t \lambda I  \right)w^t - \eta_t c \odot \text{sign}(w^t) \odot h(w^t) + \eta_t   c \odot  \text{sign}(w^t) \odot \text{vec}_j \left(\sum_{l=0}^{2m} \widetilde{\alpha}_l \mathbb{E}_{\vb x   } \left[ x_l\sigma \left(  \text{sign}(w^t_j) x_j \right)  \right] \right)   - \eta_t \Gamma_t\\
\nonumber
&= \left( I- \eta_t \lambda I  \right)w^t - \eta_t c \odot \text{sign}(w^t) \odot h(w^t) +  \eta_t   c \odot  \text{sign}(w^t) \odot {\text{sign}(w^t)  \odot \alpha \over 2} - \eta_t \Gamma_t\\
\label{eq:main_iteration_u}
&= \left( I- \eta_t \lambda I  \right)w^t - \eta_t c \odot \text{sign}(w^t) \odot h(w^t) + \eta_t {c \odot \alpha \over 2} - \eta_t \Gamma_t,
\end{align}
\normalsize
where $\alpha \in \R^{4m+1}$ with $\alpha_j = \widetilde{\alpha}_j$ for $j \in [0,2m]$, $\alpha_j = - \widetilde{\alpha}_{-j}$ for $j \in [-2m, -1]$.

\subsection{Fourier feature learning in the diagonal layer}

We employ symmetric initialization for the second layer weight $c$, which is commonly used in neural network initialization \cite{shi2023provable}: Let $(c_0)_j \sim \text{Unif}\{-r_c /\sqrt{m}, r_c /\sqrt{m} \}$ for $j \ge 0$ and $(c_0)_{-j} = -(c_0)_j$ for $j \ge 1$. Here, $r_c$ is a constant which will be determined later. Note that $\|c\|_\infty \le r_c/\sqrt{m}$ and $\|c\|_2 \le r_c$.

We have the following theorem about Fourier feature learning of the first layer. 

\begin{thm}
\label{thm:feature_learning}
Suppose that Assumptions \ref{assumption:recovery_condition} and \ref{assumption:recovery_condition2} hold for the target function $f$. 
Let $Q_1$ be any real number with $Q_1 \ge {1 \over \lambda} \left \| {c_0\odot \alpha\over 2} \right \|_\infty$ with $0 < \lambda < 1$. Set the step size $\eta_t = \eta$ such that $\eta < 1/\lambda$. Then, for any $\delta \in (0,1/T)$ with probability at least $T\delta$, we have
\footnotesize
\begin{align*}
&\left \| w^{T} - {1-(1-\eta \lambda)^T \over \lambda}  {c_0 \odot \alpha\over 2}  \right \|_\infty \\
&\le \left(1 -  \eta \lambda \right)^{T-1}  \left \|  w^1 -  \eta_1 {c_0 \odot \alpha\over 2}  \right \|_\infty + \left( \sqrt{5}r^2_c Q_1+ {2\sqrt{5} C  r^2_c Q_1\sqrt{\log m} \log \delta^{-1} \over \sqrt{B} } + {2 C (\|f\|_\infty   + \kappa) r_c\sqrt{\log m} \log \delta^{-1} \over \sqrt{B m} }  \right) \cdot  {1 \over \lambda}.
\end{align*}
\normalsize
\end{thm}

\begin{cor}
\label{cor:First_layer_alignment}
Assume the symmetric initialization for $c$, i.e.,  $c_{-i} = -c_i$ and $|c_i| = r_c/\sqrt{m}$ and suppose that $Q_1$ is of the order of $O \left({r_c \|\alpha\|_\infty \over \lambda \sqrt{m}} \right)$. For sufficiently large $T$, the second term in the approximation error in Theorem \ref{thm:feature_learning} dominates. In particular, when $r_c < 1$ and $B > 1/r^2_c$, then the error term is of the order of $O \left({r^2_c \over \lambda \sqrt{m}} \right)$. In other words, 
\begin{align*}
&\left \| w^{T} - {1-(1-\eta \lambda)^T \over \lambda}   {c_0 \odot \alpha\over 2}  \right \|_\infty \le O \left({r^2_c (\|\alpha\|_\infty + \|f\|_\infty ) \over \lambda \sqrt{m}} \right),
\end{align*}
whereas the $j$-th entry of the learned feature ${1-(1-\eta \lambda)^T \over \lambda}  {(c_0 \odot \alpha)_j \over 2}\sim O \left({r_c |\alpha_j| \over \lambda \sqrt{m}} \right)$.
\end{cor}

\begin{cor}
   Due to the symmetric initialization of $c_0$ and $\alpha_j = \widetilde{\alpha}_j, \alpha_{-j}  = - \widetilde{\alpha}_j$, we have $\text{sign}((c_0 \odot \alpha)_j) = \text{sign}((c_0 \odot \alpha)_{-j})  = \text{sign}(\widetilde{\alpha}_j)$ for $j \in [1,2m]$. Hence, Theorem \ref{thm:feature_learning} implies that after some iterations $T$, for $j$ with $|\widetilde{\alpha}_j| > 0$,  we have $\text{sign}(w^T_j) = \text{sign}(w^T_{-j})$.
\end{cor}

\begin{cor}
Set the step size $\eta = {2 \log T \over \lambda T}$ with $\eta \lambda < 1$, which is satisfied for any moderately large $T$. Since $\left(1 -  \eta \lambda \right)^T \le \exp(-\lambda \eta T) \le \exp(-2\log T) \lesssim {1 \over T}$.
\small
\begin{align*}
&\lesssim {1 \over T}  \left \|  w^1 -  \eta_1 {c_0 \odot \alpha\over 2}  \right \|_\infty + \left( r^2_c Q_1+ {2\sqrt{5} C  r^2_c Q_1\sqrt{\log m} \log \delta^{-1} \over \sqrt{B} } + {2 C (\|f\|_\infty   + \kappa) r_c\sqrt{\log m} \log \delta^{-1} \over \sqrt{B m} }  \right) \cdot  {1 \over \lambda}.
\end{align*}
\normalsize
\end{cor}

\begin{proof} [Proof of Theorem \ref{thm:feature_learning}]
We use the convention $\prod\limits_{i=p}^q = 1$ if $p > q$.
\small
\begin{align}
\nonumber
&\left \| w^{t+1} -   \sum_{p=1}^{t} \eta_p \prod_{i=p+1}^{t} \left( 1 -  \eta_i \lambda \right)  {c_0 \odot \alpha\over 2}  \right \|_\infty \\
\label{eq:main_col_bound1}
&= \left \| P_{Q_1} \left(u^t \right) - P_{Q_1} \left( \sum_{p=1}^{t} \eta_p \prod_{i=p+1}^{t} \left( 1 -  \eta_i \lambda \right)  {c_0 \odot \alpha\over 2} \right) \right \|_\infty \\
\label{eq:main_col_bound2}
&= \left \| P_{Q_1} \left(u^t -  \sum_{p=1}^{t} \eta_p \prod_{i=p+1}^{t} \left( 1 -  \eta_i \lambda \right)  {c_0 \odot \alpha\over 2}\right) \right \|_\infty \\
\label{eq:main_col_bound3}
&\le \left \| u^t - \sum_{p=1}^{t} \eta_p \prod_{i=p+1}^{t} \left( 1 -  \eta_i \lambda \right)  {c_0 \odot \alpha\over 2} \right \|_\infty \\
\label{eq:main_col_bound4}
&= \left \| \left( I- \eta_t \lambda I  \right)w^t + \eta_t c_0 \odot  \left({ \alpha \over 2} - \text{sign}(w^t) \odot h(w^t)  \right) - \eta_t \Gamma_t -  \sum_{p=1}^{t} \eta_p \prod_{i=p+1}^{t} \left( 1 -  \eta_i \lambda \right)  {c_0 \odot \alpha\over 2}\right \|_\infty \\
\label{eq:main_col_bound5}
&= \left \| \left( I- \eta_t \lambda I  \right)w^t - \eta_t c_0 \odot \text{sign}(w^t) \odot h(w^t) - \eta_t \Gamma_t - \left( I- \eta_t \lambda I  \right) \sum_{p=1}^{t-1} \eta_p \prod_{i=p+1}^{t-1} \left( 1 -  \eta_i \lambda \right)  {c_0 \odot \alpha\over 2}\right \|_\infty \\
\nonumber
&= \left \| \left( I- \eta_t \lambda I  \right) \left( w^t -  \sum_{p=1}^{t-1} \eta_p \prod_{i=p+1}^{t-1} \left( 1 -  \eta_i \lambda \right)  {c_0 \odot \alpha\over 2} \right) - \eta_t c_0 \odot \text{sign}(w^t) \odot h(w^t) - \eta_t \Gamma_t \right \|_\infty \\
\label{eq:main_col_bound6}
&\le \left \| \left( I- \eta_t \lambda I  \right) \left( w^t -  \sum_{p=1}^{t-1} \eta_p \prod_{i=p+1}^{t-1} \left( 1 -  \eta_i \lambda \right)  {c_0 \odot \alpha\over 2} \right) \right \|_\infty  + \eta_t \left \|c_0 \odot  h(w^t) \right \|_\infty  + \eta_t \left \|\Gamma_t \right \|_\infty \\
\nonumber
&\le (1- \eta_t \lambda ) \left \|  w^t -\sum_{p=1}^{t-1} \eta_p \prod_{i=p+1}^{t-1} \left( 1 -  \eta_i \lambda \right)  {c_0 \odot \alpha\over 2}  \right \|_\infty  + \eta_t \left \|c_0 \odot  h(w^t) \right \|_\infty + \eta_t \left \|\Gamma_t \right \|_\infty \\
\nonumber
&\le (1- \eta_t \lambda ) \left \|  w^t -\sum_{p=1}^{t-1} \eta_p \prod_{i=p+1}^{t-1} \left( 1 -  \eta_i \lambda \right)  {c_0 \odot \alpha\over 2}  \right \|_\infty + \eta_t \|c_0\|_\infty \|  h(w^t)  \|_\infty + \eta_t \left \|\Gamma_t \right \|_\infty\\
\label{eq:main_col_bound7}
&\le (1- \eta_t \lambda ) \left \|  w^t -\sum_{p=1}^{t-1} \eta_p \prod_{i=p+1}^{t-1} \left( 1 -  \eta_i \lambda \right)  {c_0 \odot \alpha\over 2}  \right \|_\infty +  \eta_t \|c_0\|_\infty \|c_0\|_2 \|  w^t  \|_2 + \eta_t \left \|\Gamma_t \right \|_\infty\\
\label{eq:main_col_bound8}
&\le (1- \eta_t \lambda ) \left \|  w^t -\sum_{p=1}^{t-1} \eta_p \prod_{i=p+1}^{t-1} \left( 1 -  \eta_i \lambda \right)  {c_0 \odot \alpha\over 2}  \right \|_\infty + \eta_t \|c_0\|_\infty \|c_0\|_2 \sqrt{5m}Q_1 + \eta_t \left \|\Gamma_t \right \|_\infty.
\end{align}
\normalsize
Here, the equality \eqref{eq:main_col_bound1} is from the relation $w^{t+1} = P_{Q_1} \left(u^t \right)$ and the fact that 

$Q_1 > \|\sum_{p=1}^{t} \eta_p \prod_{i=p+1}^{t} \left( 1 -  \eta_i \lambda \right)  {c_0 \odot \alpha\over 2}\|_\infty$ for all $t$ since $\eta_i = \eta$. \eqref{eq:main_col_bound2} follows from the linearity of the projection operator $P_{Q_1} $. The inequality \eqref{eq:main_col_bound3} is from the fact that the orthogonal projection $P_{Q_1}$ of a vector $u$ to the box $[-Q_1, Q_1]^{4m+1}$ does not increase the $\ell_\infty$-norm of $u$. The identity \eqref{eq:main_iteration_u} yields \eqref{eq:main_col_bound4}, which in turn gives \eqref{eq:main_col_bound5} using the convention $\prod\limits_{i=p}^q = 1$ if $p > q$.
The inequality \eqref{eq:main_col_bound6} follows from the triangle inequality. 
The inequality \eqref{eq:main_col_bound7} is from the bound for $h(w)$ in \eqref{eq:bound_h}. Lastly, \eqref{eq:main_col_bound8} is from the fact that $\|  w^t  \|_2 \le \sqrt{4m+1} \|  w^t  \|_\infty \le \sqrt{4m+1} Q_1$.

By induction on $t$, we have
\begin{align}
\nonumber
&\left \| w^{t+1} -   \sum_{p=1}^{t} \eta_p \prod_{i=p+1}^{t} \left( 1 -  \eta_i \lambda \right)  {c_0 \odot \alpha\over 2}  \right \|_\infty\\
\nonumber
&\quad \le \prod_{p=1}^t\left(1 -  \eta_p \lambda \right)  \left \| w^1 - \eta_1 {c_0 \odot \alpha\over 2}  \right \|_\infty \\
\label{eq:l2_norm_layer1_bound1}
&\qquad + \|c_0\|_\infty \|c_0\|_2 \sqrt{5m}Q_1 \cdot \sum_{p=1}^{t} \eta_p \prod_{i=p+1}^{t} \left( 1 -  \eta_i \lambda \right) +  \sum_{p=1}^{t} \eta_p  \|\Gamma_p \|_\infty  \prod_{i=p+1}^{t} \left( 1 -  \eta_i \lambda \right).
\end{align}

Let $B$ be the size of the minibatch of Algorithm \ref{alg:layerwise-training}. 
Then, $\Gamma_t = (\Gamma_t^{(1)} + \Gamma_t^{(2)} + \dots + \Gamma_t^{(B)})/B$, 
where each $\Gamma_t^{(i)}$ is a bounded independent mean-zero sub-Gaussian random vector with the bound given by as follows. 

Let $|z| \le \kappa$.
\begin{align*}
\left \|\Gamma_t^{(i)} \right \|_\infty
&\le 2\left | c_0^T  \sigma(w  \odot \vb x)  - f(\vb x)  - z \right | \|c_0  \odot \vb x\|_\infty \\
&\le 2\left | c_0^T  \sigma(w  \odot \vb x)  - f(\vb x)  - z \right | \|c_0\|_\infty  \\
&\le 2 \left(  | c_0^T  \sigma(w  \odot \vb x)| +|f(\vb x)|  + |z|   \right) \|c_0\|_\infty  \\
&\le 2 \left(  \|c_0\|_2  \| \sigma(w  \odot \vb x) \|_2 + \|f\|_\infty  + |z|  \right) \|c\|_\infty \\
&\le 2 \left(  \|c_0\|_2  \| w \|_2 + \|f\|_\infty  + |z|  \right) \|c_0\|_\infty \\
&\le 2 \left( \sqrt{5m}Q_1 \|c_0\|_2  + \|f\|_\infty  + \kappa \right) \|c_0\|_\infty := M
\end{align*}
where we used the fact that $\|w^t \|_2 \le \sqrt{4m+1} \|w^t \|_\infty =  \sqrt{4m+1} \|P(u^{t-1}) \|_\infty \le \sqrt{4m+1} Q_1$ and the fact that $|\cos (\pi a_{2i+1} \theta )| \le 1$ and $|\sin (\pi a_{2j} \theta)| \le 1$.

Note that since $\|c_0\|_2 \le \sqrt{4m+1} \|c_0\|_\infty $ and from our assumption on $c_0$, we have
\[
M  \le 2 \left( \sqrt{5m}Q_1 r_c  + \|f\|_\infty  + \kappa \right) \|c\|_\infty  \le 2( \sqrt{5m}r_c Q_1 + \|f\|_\infty   + \kappa) \cdot {r_c \over \sqrt{m}}.
\]

Recall that $\Gamma_t^{(1)} + \Gamma_t^{(2)} + \dots + \Gamma_t^{(B)}$ is the sum of bounded independent mean-zero random variables with the bound $\left \|\Gamma_t^{(i)} \right \|_\infty \le M$.

Hence, each component of the vector $\Gamma_t^{(1)} + \Gamma_t^{(2)} + \dots + \Gamma_t^{(B)}$, say $j$-th component of $[\Gamma_t]_j$, where $[\Gamma_t]_j= ([\Gamma_t^{(1)}]_{j} + [\Gamma_t^{(2)}]_{j} + \dots + [\Gamma_t^{(B)}]_{j})/B$ is the sum of bounded independent mean-zero sub-Gaussian scalar random variables with the bound $M$.

From Hoeffding's inequality \cite{vershynin2018high},
\[
\mathbb{P}(\left|[\Gamma_t]_j \right| > \epsilon M) = \mathbb{P}(\left|[\Gamma_t]_j- \mathbb{E}[\Gamma_t]_j \right| > \epsilon M) = \exp( -2 B \epsilon^2).
\]
Taking the union bound over all the components 
\[
\mathbb{P}(\|\Gamma_t\|_\infty > \epsilon M) \le (4m+1)  \exp(-2 B \epsilon^2 ) \le 5 \exp( \log m -2 B \epsilon^2 ).
\]
This implies that $\|\Gamma_t\|_\infty \lesssim M {\sqrt{\log m} \log \delta^{-1} \over \sqrt{B}}$ with probability at least $\delta$.


Using this bound on the inequality \eqref{eq:l2_norm_layer1_bound1}, we have
\begin{align*}
&\left \| w^{t+1} -  \sum_{p=1}^{t} \eta_p \prod_{i=p+1}^{t} \left( 1 -  \eta_i \lambda \right)  {c_0 \odot \alpha\over 2}  \right \|_\infty \\
&\le \prod_{p=1}^t\left(1 -  \eta_p \lambda \right)  \left \|  w^1 -  \eta_1 {c_0 \odot \alpha\over 2}  \right \|_\infty  \\
&\qquad + \left( \|c_0\|_\infty \|c_0\|_2 \sqrt{5m}Q_1 + {CM \sqrt{\log m} \log \delta^{-1} \over \sqrt{B }}  \right) \cdot \sum_{p=1}^{t} \eta_p \prod_{i=p+1}^{t} \left( 1 -  \eta_i \lambda \right).
\end{align*}

In particular, for the constant step size $\eta_t = \eta$, the bound above implies that 
\small
\begin{align*}
&\left \| w^{t+1} -   \sum_{p=1}^{t} \eta_p \prod_{i=p+1}^{t} \left( 1 -  \eta_i \lambda \right)  {c_0 \odot \alpha\over 2}  \right \|_\infty\\
&\le \left(1 -  \eta \lambda \right)^t  \left \|  w^1 -  \eta_1 {c_0 \odot \alpha\over 2}  \right \|_\infty \\
&\qquad + \left( \|c\|_\infty \|c\|_2 \sqrt{5m}Q_1 + {CM\sqrt{\log m} \log \delta^{-1} \over \sqrt{B}}  \right) \cdot  \eta   \sum_{p=1}^{t} \left( 1 -  \eta \lambda \right)^{t-p}\\
&\le \left(1 -  \eta \lambda \right)^t  \left \|  w^1 -  \eta_1 {c_0 \odot \alpha\over 2}  \right \|_\infty + \left( \|c_0\|_\infty \|c_0\|_2 \sqrt{5m}Q_1 + {CM\sqrt{\log m} \log \delta^{-1} \over \sqrt{B}}  \right) \cdot  {1 \over \lambda} \\
&\le \left(1 -  \eta \lambda \right)^t  \left \|  w^1 -  \eta_1 {c_0 \odot \alpha\over 2}  \right \|_\infty + \left( {r_c \over \sqrt{m}} r_c \cdot \sqrt{5m}Q_1 +  {2 C (\sqrt{5m}r_c Q_1+ \|f\|_\infty  + \kappa) r_c\sqrt{\log m} \log \delta^{-1} \over \sqrt{B m } }  \right) \cdot  {1 \over \lambda}\\
&\le \left(1 -  \eta \lambda \right)^t  \left \|  w^1 -  \eta_1 {c_0 \odot \alpha\over 2}  \right \|_\infty \\
&\qquad + \left( \sqrt{5}r^2_c Q_1+ {2\sqrt{5} C  r^2_c Q_1\sqrt{\log m} \log \delta^{-1} \over \sqrt{B} } + {2 C (\|f\|_\infty  + \kappa) r_c\sqrt{\log m} \log \delta^{-1} \over \sqrt{B m} }  \right) \cdot  {1 \over \lambda},
\end{align*} 
\normalsize
with probability at least $t \delta$ by the union bound. 
\end{proof}

\section{Approximation of differentiable periodic mixtures}

Assume that the target function $f$ is of the form of the $k$ mixtures of $r$-times continuously differentiable and periodic functions $g_i( \cos ( a_{2i+1} x))$ and $h_j( \sin ( a_{2j} x))$ with link functions $g_i$ and $h_j$. Also, suppose that  $f$ satisfies the recovery conditions in Assumptions \ref{assumption:recovery_condition} and \ref{assumption:recovery_condition2}. 

Then, from Bernstein's inequality for approximation of $r$-times differentiable functions by the sine-cosine (Fourier) expansion \cite{stein2011fourier, katznelson2004introduction}, each $g_l( \cos ( a_{2l+1} x))$ can be approximated by the expansion as below. Let $J_{2l+1}$ be the set of nonzero Fourier coefficients of $g_l( \cos ( a_{2l+1} x))$ in $[0, 2m]$. 
\begin{align*}
&\left| \sum_{i \in J_{2l+1}}  \alpha_{2i+1}\cos ( i \theta) -  g_l( \cos ( a_{2l+1} \theta)) \right| \\
&=\left| \sum_{i \in J_{2l+1}}  \alpha_{2i+1} \vb x_{2i+1} -  g_l( \cos ( a_{2l+1} \theta)) \right| \\
&= \left| \sum_{i \in J_{2l+1}}  \alpha_{2i+1} \left[ \sigma({\vb x_{2i+1})+ \sigma( - \vb x_{2i+1}}) \right]- g_l( \cos ( a_{2l+1} x)) \right| \\
&= \left| \sum_{i \in J_{2l+1}}  \alpha_{2i+1} \left[ \sigma(\text{sign} ( (c_0 \odot \alpha)_{2i+1} )  {\vb x_{2i+1} )+ \sigma(- \text{sign} ( (c_0 \odot \alpha)_{-(2i+1)} ) \vb x_{2i+1}}) \right]- g_l( \cos ( a_{2l+1} x)) \right| \\
&= \left| \sum_{i \in J_{2l+1}}  \alpha_{2i+1} \left[ \sigma(\text{sign} ( (c_0 \odot \alpha)_{2i+1} )  {\vb x_{2i+1} )+ \sigma(\text{sign} ( (c_0 \odot \alpha)_{-(2i+1)} ) \vb x_{-(2i+1)}}) \right]- g_l( \cos ( a_{2l+1} x)) \right| \\
&= \left| \sum_{i \in J_{2l+1}}  \alpha_{2i+1} \left[ \sigma(\text{sign} ( w^T_{2i+1} )  {\vb x_{2i+1} )+ \sigma(\text{sign} ( w^T_{-(2i+1)} ) \vb x_{-(2i+1)}}) \right]- g_l( \cos ( a_{2l+1} x)) \right| \\
&= \left| \sum_{i \in J_{2l+1}}  \alpha_{2i+1} \left[ \sigma({w^T_{2i+1} \over |w^T_{2i+1}| } \vb x_{2i+1} )+ \sigma( {w^T_{-(2i+1)} \over |w^T_{-(2i+1)}| }  \vb x_{-(2i+1)}) \right]- g_l( \cos ( a_{2l+1} x)) \right| \\
&= \left| \sum_{i \in J_{2l+1}}  \alpha_{2i+1} \left[  {1 \over |w^T_{2i+1}|} \sigma({w^T_{2i+1}  } \vb x_{2i+1} )+ {1 \over |w^T_{-(2i+1)}| } \sigma( {w^T_{-(2i+1)}}  \vb x_{-(2i+1)}) \right]- g_l( \cos ( a_{2l+1} x)) \right| \\
&\lesssim {C(g_l) \over m^r}, 
\end{align*}
where the third identity is from $\text{sign} ( (c_0 \odot \alpha)_{2i+1} ) = \text{sign} ( (c_0 \odot \alpha)_{-(2i+1)} )$ (due to the symmetric initialization for $(c_0)_{-j} = - (c_0)_{j} $ and $\alpha_{-j} = - \alpha_j$) and the fifth identity is from 
\small
\begin{align*}
&\left| w^T_j - \eta \sum_{p=1}^{T-1} \left( 1 -  \eta_i \lambda \right)^{T-p-1} {(c_0 \odot \alpha)_j \over 2} \right| \\
&\le \left(1 -  \eta \lambda \right)^{T-1}  \left \|  w^1 -  \eta_1 {c_0 \odot \alpha\over 2}  \right \|_\infty \\
&\qquad + \left( \sqrt{5}r^2_c Q_1+ {2\sqrt{5} C  r^2_c Q_1\sqrt{\log m} \log \delta^{-1} \over \sqrt{B} } + {2 C (\|f\|_\infty  + \kappa) r_c\sqrt{\log m} \log \delta^{-1} \over \sqrt{B m} }  \right) \cdot  {1 \over \lambda} \\
&\le O \left({r^2_c \|\alpha\|_\infty \over \lambda \sqrt{m}} \right).   
\end{align*}
\normalsize
for sufficiently large $T$. Thus, ${|w^T_i|} \ge { |(c_0 \odot \alpha)_i| \over 4 \lambda} - O \left({r^2_c (\|\alpha\|_\infty + \|f\|_\infty )  \over \lambda \sqrt{m}} \right) =  { r_c  |\alpha_i| \over 4  \lambda \sqrt{m}} - O \left({r^2_c (\|\alpha\|_\infty + \|f\|_\infty )  \over \lambda \sqrt{m}} \right)$. 
From the assumption on $\min\limits_{i: \alpha_i  \neq 0} |\alpha_i| > O \left(8{r_c} (\|\alpha\|_\infty + \|f\|_\infty ) \right)$, ${|w^T_i|} \ge  O \left( {r^2_c (\|\alpha\|_\infty + \|f\|_\infty )  \over 8 \lambda \sqrt{m}} \right)$
or ${1 \over |w^T_i|} \le {8 \lambda \sqrt{m} \over  r^2_c (\|\alpha\|_\infty + \|f\|_\infty )  }  $ for $i \in J_{2l+1}$. If we set $\widetilde{c}_i = {\alpha_i \over |w^T_i|}$, then $\|\widetilde{c}\|_\infty \le   {8 \|\alpha\|_\infty \lambda \sqrt{m} \over  r^2_c (\|\alpha\|_\infty + \|f\|_\infty ) } \le {8 \lambda \sqrt{m} \over  r^2_c}$. 

Hence, there exists $\widetilde{c}_{2i+1}$ such that 
\begin{align*}
&\left| \sum_{i=-m}^m [\widetilde{c}_{2i+1} \sigma(w^T_{2i+1} \vb x_{2i+1}) - g_l( \cos ( a_{2l+1} x)) \right| \lesssim {C(g_l) \over m^r}
\end{align*} where $|\widetilde{c}_i| \le {8 \lambda \sqrt{m} \over  r^2_c } $.

For example, when the link function $g_l$ is a degree $p$ polynomial, a simple calculation from the definition of total variation of a function gives
\[
C(g_l) \le  |a_{2l+1}| \max\limits_{w \in [-1,1]}|g_l'(w)| (1 + 2+ \dots + p) \lesssim p^2 |a_{2l+1}| \max\limits_{w \in [-1,1]}|g_l'(w)|.   
\]
Similarly, from the approximation of $r$-times differentiable odd functions by Fourier series, for each $h_l$, there exists $\widetilde{c}_{2i}$ such that 
\[
\left| \sum_{i=-m}^m \widetilde{c}_{2i} \sigma(w^T_i \vb x_i) - h_l( \sin ( a_{2l} x)) \right| \lesssim {C(h_l) \over m^r}, 
\] for all $x \in [-1,1]$, where $C(h_l)$ is the total variation of $h_l(  \sin ( a_{2l+1} x))$. 
Again, when the link function $h_l$ is a degree $q$ polynomial, we have
\[
C(h_l)  \lesssim q^2 |a_{2l}| \max\limits_{w \in [-1,1]}|h_l'(w)|.   
\]

Hence, using the decomposition $f = f_{\text{even}} + f_{\text{odd}}$ and  by the triangle inequality, we have
\begin{align}
\label{eq:approximation_bound_polynomial}
\left| \sum_{i=-2m}^{2m} \widetilde{c}_i \sigma(w^T_i \vb x_i) - f(\vb x) \right| \lesssim {\sum_{l=1}^s C(g_l) + \sum_{l=1}^{s'} C(h_l) \over m^r}, 
\end{align}
for all $x \in [-1,1]$ with $|\widetilde{c}_i| \le {8 \lambda \sqrt{m} \over  r^2_c }$ for $\alpha_i \neq 0$ and $\widetilde{c}_i = 0$ if $\alpha_i = 0$.

\begin{lem}
\label{lem:empirical_loss_optimal_c}
Suppose that Assumptions \ref{assumption:recovery_condition} and \ref{assumption:recovery_condition2} hold for the target function $f$. 
Let $T$ be the number of iterations and $B$ is the mini-batch size used in training the first layer. Set $L = TB$. Suppose that
$\min\limits_{i: \alpha_i \neq 0} |\alpha_i| > O \left(8{r_c} (\|\alpha\|_\infty + \|f\|_\infty ) \right)$ and 
\footnotesize
\begin{align*}
&\left(1 -  \eta \lambda \right)^{T-1}  \left \|  w^1 -  \eta_1 {c_0 \odot \alpha\over 2}  \right \|_\infty + \left( \sqrt{5}r^2_c Q_1+ {2\sqrt{5} C  r^2_c Q_1\sqrt{\log m} \log \delta^{-1} \over \sqrt{B} } + {2 C (\|f\|_\infty  + \kappa) r_c\sqrt{\log m} \log \delta^{-1} \over \sqrt{B m} }  \right) \cdot  {1 \over \lambda} \\
&\le O \left({r^2_c \|\alpha\|_\infty \over \lambda \sqrt{m}} \right).
\end{align*}
\normalsize

Then, there exists a second layer weight $\widetilde{c}$ such that the empirical risk of $\hat{R}(w^T, c)$ satisfies
\begin{align*}
&\hat{R}(w^T, \widetilde{c}) \\
&= {1 \over L} \sum_{i=1}^L \ell (\hat{f}(\vb x^{(i)}; w^T, \widetilde{c}) , y^i) \\
&= {1 \over L} \sum_{i=1}^L \left(
\widetilde{c}^\top \widetilde{\sigma}(w^T \odot \vb x^{(i)}) - y^i \right)^2 \\
& \lesssim  {(\sum_{l=1}^s C(g_l) + \sum_{l=1}^{s'} C(h_l))^2 \over m^{2r}} + \sqrt{\nu \over L} + 2\mathbb{E}[\epsilon^2].
\end{align*}    
\end{lem}

\begin{proof}
Consider the empirical risk of $\hat{R}(w^T, c)$. 
\[
\hat{R}(w^T, c) = {1 \over L} \sum_{i=1}^L \ell (\hat{f}(\vb x^{(i)}; w^T, \widetilde{c}) , y^i) = {1 \over L} \sum_{i=1}^L (\hat{f}(\vb x^{(i)}; w^T, \widetilde{c})  - f( \vb x_i) - \epsilon^i)^2. 
\]

From the inequality $(a+b)^2 \le 2a^2 + 2b^2$, we have
\begin{align*}
    &(\hat{f}(\vb x_i; w^T, \widetilde{c})  - f(\vb x_i) - \epsilon^i)^2\\ 
    &\le   2( \hat{f}(\vb x_i; w^T, \widetilde{c}) -  f( \vb x_i)   )^2 + 2(\epsilon^i)^2\\
    &\le  2{(\sum_{l=1}^s C(g_l) + \sum_{l=1}^{s'} C(h_l))^2 \over m^{2r}} + 2(\epsilon^i)^2
\end{align*}

Because $\epsilon^i$ are i.i.d sub-Gaussian, $(\epsilon^i)^2$ are i.i.d. sub-exponential random variables. By the concentration inequality of the sub-exponential random variable with sub-exponential norm $\nu$ \cite{vershynin2018high}, we have
\[
{1 \over L} \sum_{i=1}^L (\epsilon^i)^2 - \mathbb{E}[\epsilon^2] \lesssim \sqrt{\nu \over L}.
\]
\end{proof}

By slightly abusing the notation, we also denote the empirical loss function $\hat{R}(w^T, c)$ in $c$ as $\hat{R}(c)$. Let $\hat{c}$ be the minimizer of the following optimization problem:
\begin{align}
\label{eq:2nd_layer_optimization}
\min\limits_{c: \|c\|_\infty \le O({ 8 \lambda \sqrt{m} / r^2_c })} \hat{R}(c).    
\end{align}

\section{Training the second layer}
Suppose that the training of the first layer was carried out using the sample $\{\vb x^{(i)}, y^i \}_{i=1}^L$. We will train the second layer weights $c$ using another SGD with the same sample $\{\vb x^{(i)}, y^i \}_{i=1}^L$. 

Let $e_c^t = \nabla_c \ell (f(\vb x; w^T, c), y) - \mathbb{E}_{\vb x, y} [\nabla_c \ell (f(\vb x; w^T, c), y)]$ be the stochastic noise of the gradient with respect to $c$ at the $t$-th iteration. 

We first estimate the Hessian of the regularized empirical loss $\hat{R}(c)$ with respect to $c$. For any unit vector $v$, we have
\begin{align*}
    \inner{v, \nabla^2_c \hat{R}(c)  v}
    &= \inner{v,  {1 \over L} \sum_{i=1}^L \nabla^2_c \ell (\hat{f}(\vb x^{(i)}; w^T, c) , y^i)v }  \\
    &= {1 \over L} \sum_{i=1}^L  \inner{v,  \sigma(w^T \odot \vb x^{(i)}) \sigma(w^T \odot \vb x^{(i)})^\top v } \\
    &\le {1 \over L} \sum_{i=1}^L \|\sigma(w^T \odot \vb x^{(i)})\|_2^2 \\
    &\le {1 \over L} \sum_{i=1}^L \|w^T \odot \vb x^{(i)}\|_2^2 \\
    &\le {1 \over L} \sum_{i=1}^L \|w^T \|_2^2\\
    &\le 5m{Q^2_1},
\end{align*} where the first inequality is from Cauchy-Schwarz inequality, the third  is from the fact that each component of $\vb x^{(i)}$ are sinusoids, so their magnitudes are bounded by $1$, and the last inequality is from the fact that $\|w^t\|_2 \le \sqrt{5m} Q_1$ for all $t$.

Set $Q_2 = O({ 8 \lambda \sqrt{m} / r^2_c })$ in the projected SGD in Algorithm \ref{alg:layerwise-training}.
Then, 
\begin{align*}
\|e_c^t\|_2 &\le 2 \|\nabla_c l\|_2 \\
&\le 2\left | c^\top  \sigma(w^T  \odot \vb x)  - f(\vb x)  - z \right | \|w^T  \odot \vb x\|_2 \\
&\le 2 \left( \|c\|_\infty \|\sigma(w^T  \odot \vb x)\|_1 + \|f\|_\infty  + \kappa \right) \|w^T\|_2 \\
&\le 2 \left( Q_2 \|w^T\|_1  + \|f\|_\infty  + \kappa \right) \|w^T\|_2 \\
&\le 2 \left( 5m Q_1 Q_2  + \|f\|_\infty  + \kappa \right) \cdot \sqrt{5m} Q_1\\
&= O \left( m + \|f\|_\infty  + \kappa \right) \cdot {r_c \|\alpha\|_\infty /\lambda}\\
& := M', 
\end{align*}  

First, recall that $f(\vb x; w^T, c) = c^\top \sigma(w^T  \odot \vb x) = \sigma(w^T \odot \vb x)^\top c$.  The stochastic gradient of the regularized loss function with respect to $c$
\begin{align*} 
        &\nabla_c \ell (f(\vb x; w^T, c), y) \\
        &= (\hat{y} - y) \cdot  \sigma(w^T \odot \vb x)  \\
        &= (\sigma(w^T  \odot \vb x)^\top c - y) \cdot  \sigma(w^T  \odot \vb x),
\end{align*}
and 
\[
\nabla^2_c \ell (f(\vb x; w^T, c), y) = \sigma(w^T \odot \vb x) \sigma(w^T \odot \vb x)^\top.
\]

Note also that the Hessian $\nabla^2_c \ell (f(\vb x; w^T, c), y)$ is a constant in $c$ and $\nabla^2_c \ell (f(\vb x; w^T, c), y) \preccurlyeq 5m Q^2_1 \cdot I$.

Since $c \rightarrow \ell(w^T, c)$ is a convex function in $c$ and the Hessian $\nabla^2_c$ is bounded, there are standard results in optimization about its convergence. In particular, we will apply the following consequence of Theorem 3.3 in \cite{liu2023revisiting} about the high probability bound of the last iterate of the projected SGD. 

\begin{thm}
\label{thm:convergence_2nd_layer}
    Let $\delta \in (0,1)$. 
    Let $\eta'_t = \min \left\{ {1 \over 10 Q^2_1}, {\eta' \over \sqrt{T'}} \right\}$ with $\eta' = \Theta \left( \sqrt{ \|\hat{c} - c_0\|_2^2 \over M' \log(1/\delta) \log T' } \right)$. 
    Then, with probability at least $1 - \delta$, we have
    \begin{align*}
        \hat{R}(c^{T'+1}) - \hat{R}(\hat{c}) 
        &\le O \left({M'\|\hat{c} - c_0\|_2^2  \over T'} + {\sqrt{M' \log(1/\delta) \|\hat{c} - c_0\|_2^2 \log T'} \over \sqrt{T'} }   \right).
    \end{align*}  
\end{thm}

\begin{cor}
Under the same condition in Theorem \ref{thm:convergence_2nd_layer},
    \begin{align*}
        & \hat{R}(c^{T'+1}) - \hat{R}(\hat{c}) \\
        &\le O \left({M'  O( \lambda^2 m^2 / r^4_c + r^2_c) \over T'} + {\sqrt{M' \log(1/\delta)  O( \lambda^2 m^2 / r^4_c + r^2_c) \log T'} \over \sqrt{T'} }   \right).
    \end{align*}    
\end{cor}
\begin{proof}
Since $\hat{c}$ is the minimizer of \ref{eq:2nd_layer_optimization}, 
$\|\hat{c}\|_\infty \le O(  8 \lambda \sqrt{m} / r^2_c )$, $\|\hat{c}\|_2^2 \le O(  \lambda^2 m^2 / r^4_c )$. Thus, $\|\hat{c} - c_0\|_2^2 \le O( \lambda^2 m^2 / r^4_c + r^2_c)$
\end{proof}

\subsection{Generalization gap}
We start with a modification of Lemma 17 in \cite{mousavi2022neural}.
First, let $R_{\tau}$ and $\hat{R}_{\tau}$ be a truncated loss function of the population risk and empirical function respectively. 

Recall that $\|w^T\|_\infty \le Q_1$ and $|z| \le \kappa$.

We define a set $S := \{c: \|c\|_\infty \le Q_2 \}$ and a hypothesis class $\mathcal{G}$ by
\[
\mathcal{G} = \{(\vb x, y) \rightarrow \ell(\hat{y}(\vb x; w^T, c),y) \wedge \tau : \|c\|_\infty \le Q_2\}, 
\] where $Q_2 = O({ 8 \lambda \sqrt{m} / r^2_c })$. 
Then, 
\[
\mathbb{E} \left[ \sup_{c \in S} R_{\tau}(w^T,c) - \hat{R}_{\tau}(w^T,c) \right] \le 2\mathcal{R}(\mathcal{G}),
\] where $\mathcal{R}(\mathcal{G})$ is the Rademacher complexity of the function class $\mathcal{G}$.

The following lemma provides the bound on $\mathcal{R}(\mathcal{G})$.
\begin{lem}
\label{lem:Complexity_class} 
Let $\tau > 1$. Then, we have
    \[
    \mathcal{R}(\mathcal{G}) \lesssim \sqrt{2\tau} \cdot {Q_2Q_1 m  \over \sqrt{L} } \le O \left(\sqrt{\tau} \|\alpha\|_\infty  \cdot  {m  \over \sqrt{L} } \right).
    \]
\end{lem}
\begin{proof}
    Let $\mathcal{F} = \{(\vb x, y) \rightarrow f_{c,w}(\vb x,y) : c \in S\}$ for $ f_{c,w}(\vb x,y) = \hat{y}(\vb x; w, c) - y$. Define the truncated loss function $h(z) = \ell(z) \wedge \tau$. Then, $h$ is $\sqrt{2 \tau}$-Lipschitz. By Talagrand's lemma \cite{mohri2018foundations}, we have $\mathcal{R}(\mathcal{G}) \le \sqrt{2 \tau} \mathcal{R}(\mathcal{F})$.
    Let $\{\xi_i\}_{i=1}^L$ be i.i.d. Rademacher random variables. Then, we have
    \begin{align*}
        \mathcal{R}(\mathcal{F}) 
        &= \mathbb{E} \left[ \sup\limits_{c \in S} {1 \over L} \sum_{i=1}^L \xi_i \left (c^T \sigma(w^T \odot \vb x^{(i)}) - y^{(i)} \right) \right] \\
        &=\mathbb{E} \left[ \sup\limits_{c \in S} {1 \over L} \sum_{i=1}^L \xi_i c^T \sigma(w^T \odot \vb x^{(i)})   \right] \\
        &=\mathbb{E} \left[ \sup\limits_{c \in S} {1 \over L} c^T \left[\sum_{i=1}^L \xi_i  \sigma(w^T \odot \vb x^{(i)}) \right]  \right] \\
        &\le \mathbb{E} \left[ \sup\limits_{c \in S} {1 \over L} \|c\|_\infty  \left \|\sum_{i=1}^L \xi_i  \sigma(w^T \odot \vb x^{(i)} )\right\|_1  \right] \\
        &\le {Q_2 \over L} \mathbb{E} \left[ \sup\limits_{c \in S}  \left \|\sum_{i=1}^L \xi_i \sigma(w^T \odot \vb x^{(i)} )\right\|_1 \right] \\
        &= {Q_2\over L} \mathbb{E} \left[  \sum_{j=-2m}^{2m} \left |\sum_{i=1}^L \xi_i \sigma(w^T_j x^{(i)}_j) \right| \right] \\
        &\le {Q_2\over L} \sum_{j=-2m}^{2m} \mathbb{E} \left[   \left |\sum_{i=1}^L \xi_i \sigma(w^T_j x^{(i)}_j) \right| \right] \\
        &\lesssim {Q_2Q_1 m \over L} \sqrt{L} \\
        &\lesssim {Q_2Q_1 m  \over \sqrt{L} },
    \end{align*}
Here, the first inequality is by H\"older's inequality. To obtain the second last inequality, first note that $|\sigma(w^T_j x^{(i)}_j)| \le |w^T_j x^{(i)}_j| \le Q_1 \cdot 1$ since $x^{(i)}_j$ are sinusoidals. Because the i.i.d. sequence $\xi_i$ and $\sigma(w^T_j x^{(i)}_j)$ are independent, this makes $\sum_{i=1}^L \xi_i \sigma(w^T_j x^{(i)}_j)$ the sum of mean-zero independent sub-Gaussian random variables with sub-Gaussian norm bounded by $Q_1$ up to a universal constant. Thus, $\left |\sum_{i=1}^L \xi_i \sigma(w^T_j x^{(i)}_j) \right|$ is a sub-Gaussian random variable with sub-Gaussian norm of the order of $O(Q_1 \sqrt{L})$ \cite{vershynin2018high}. By a well-known fact on the expectation of the sub-Gaussian random variables, we have the second last inequality. Since $Q_1 = O \left({r^2_c \|\alpha\|_\infty \over \lambda \sqrt{m}} \right)$ and $Q_2= O({8 \lambda \sqrt{m} / r^2_c })$, we have the lemma. 
\end{proof}

Note that $\mathcal{G} = \{(\vb x, y) \rightarrow \ell(\hat{y}(\vb x; w, c),y) \wedge \tau :  \|c\|_\infty \le Q_2\}$,  where $Q_1 =  O \left({r^2_c \|\alpha\|_\infty \over \lambda \sqrt{m}} \right)$ and $Q_2= O({8 \lambda \sqrt{m} / r^2_c })$. Hence, in particular, if we set $\tau  = \left( 5m Q_1 Q_2 + \|f\|_\infty  + \kappa \right)^2 = \left( \|\alpha\|_\infty + \|f\|_\infty  + \kappa  \right)^2$, $\ell(\hat{y}(\vb x; w, c),y) \wedge \tau = \ell(\hat{y}(\vb x; w, c),y)$ for all $c \in S$ because
\begin{align*}
\ell(\hat{y}(\vb x; w, c),y) 
&= ( c^T  \sigma(w  \odot \vb x)  - f(\vb x)  - z )^2 \\
&\le  \left(  \|c\|_\infty  \| \sigma(w  \odot \vb x) \|_1 + \|f\|_\infty  + |z|  \right)^2 \\
&\le  \left(  \|c\|_\infty  \| w \|_1 + \|f\|_\infty  + |z|  \right)^2 \\
&\lesssim  \left( 5m Q_1 Q_2+ \|f\|_\infty  + \kappa  \right)^2\\
&\le  O\left( m \|\alpha\|_\infty  + \|f\|_\infty  + \kappa  \right)^2.
\end{align*}

\begin{rem}
    Note that Lemma \ref{lem:Complexity_class} and the above argument on the choice of $\tau$ show that the Rademacher complexity of the function class $\mathcal{G}$ is bounded by $O\left( \left( m \|\alpha\|_\infty  + \|f\|_\infty  + \kappa  \right)  { m \over \sqrt{L} } \right)$, which is invariant in $r_c$, the parameter controls the magnitudes of the initial weights $c_0$ in the second layer (Recall that $\|c_0\|_\infty \le {r_c \over \sqrt{m}}$).
\end{rem}

The following is our main theorem showing that running Algorithm \ref{alg:layerwise-training} for $T+T'$ iterations can learn the noisy mixture of nonlinear periodic functions with a small generalization error. 
\begin{thm}
\label{thm:main_theorem1}
Suppose that Assumptions \ref{assumption:recovery_condition} and \ref{assumption:recovery_condition2} hold for the target function $f$. 
Assume the symmetric initialization for $c_0$, i.e.,  $c_{0,-i} = -c_{0,i}$ and $|c_{0,i}| = r_c/\sqrt{m}$.
Let $Q_1 \ge {1 \over \lambda} \left \| {c_0\odot \alpha\over 2} \right \|_\infty$ with $0 < \lambda < 1$ and set the step size $\eta_t = \eta$ such that $\eta < 1/\lambda$.  Suppose that $\min\limits_{i: \alpha_i \neq 0} |\alpha_i| > O \left(8{r_c} (\|\alpha\|_\infty + \|f\|_\infty ) \right)$ and that the mini-batch size $B$ satisfies $B > 1/r^2_c > 1$. Let $T$ be the number of iterations of the first phase in Algorithm \ref{alg:layerwise-training} according to Theorem \ref{thm:feature_learning}.

Then, the population loss $R(w^T, c^{T'})$  after we run Algorithm \ref{alg:layerwise-training} for $T+T'$ iterations 
\normalsize
satisfies
\begin{align*}
&R(w^T, c^{T'})   \\
&\le O \left({M'\|\hat{c} - c_0\|_2^2  \over T'} + {\sqrt{M' \log(1/\delta) \|\hat{c} - c_0\|_2^2 \log T'} \over \sqrt{T'} }   \right) +  {(\sum_{l=1}^s C(g_l) + \sum_{l=1}^{s'} C(h_l))^2 \over m^{2r}} \\
&\qquad + \sqrt{\nu \over TB} + 2\mathbb{E}[\epsilon^2]
 +  O \left( \left( m \|\alpha\|_\infty + \|f\|_\infty  + \kappa  \right) \|\alpha\|_\infty \cdot  {m  \over \sqrt{TB} } \right),
\end{align*} with probability at least $1 - (T+1)\delta$.
\end{thm}

\begin{proof}
\begin{align}
\nonumber
&R(w^T, c^{T'})   \\
\label{eq:ineq1_main_theorem}
&\le   \hat{R}(w^T, c^{T'}) +  O \left(\sqrt{\tau} \|f\|_2  \cdot  {m  \over \sqrt{L} } \right)\\
\nonumber
&=   \hat{R}(w^T, c^{T'}) -  \hat{R}(w^T, \hat{c}) +   \hat{R}(w^T, \hat{c})  +   O \left(\sqrt{\tau} \|f\|_2  \cdot  {m  \over \sqrt{L} } \right)\\
\label{eq:ineq2_main_theorem}
&\le   \hat{R}(w^T, c^{T'}) -  \hat{R}(w^T, \hat{c}) +  \hat{R}(w^T, \widetilde{c})  +   O \left(\sqrt{\tau} \|f\|_2  \cdot  {m  \over \sqrt{L} } \right)\\
\label{eq:ineq3_main_theorem}
&\le O \left({M'\|\hat{c} - c_0\|_2^2  \over T'} + {\sqrt{M' \log(1/\delta) \|\hat{c} - c_0\|_2^2 \log T'} \over \sqrt{T'} }   \right)  +   \hat{R}(w^T, \widetilde{c}) +  O \left(\sqrt{\tau} \|f\|_2  \cdot  {m  \over \sqrt{L} } \right)\\
\label{eq:ineq4_main_theorem}
&\le O \left({M'\|\hat{c} - c_0\|_2^2  \over T'} + {\sqrt{M' \log(1/\delta) \|\hat{c} - c_0\|_2^2 \log T'} \over \sqrt{T'} }   \right) +  {(\sum_{l=1}^s C(g_l) + \sum_{l=1}^{s'} C(h_l))^2 \over m^{2r}} \\
\nonumber
&\qquad + \sqrt{\nu \over L} + 2\mathbb{E}[\epsilon^2]
 +  O \left(\sqrt{\tau} \|f\|_2  \cdot  {m  \over \sqrt{L} } \right).
\end{align}
In the string of inequalities above, the inequality \eqref{eq:ineq1_main_theorem} follows from Lemma \ref{lem:Complexity_class}. The inequality \eqref{eq:ineq2_main_theorem} follows because $\hat{c}$ is the minimizer of the optimization problem \eqref{eq:2nd_layer_optimization} and $\widetilde{c}$ in Lemma \ref{lem:empirical_loss_optimal_c} 
\begin{figure}[h!]
\begin{subfigure}[b]{0.53\textwidth}
\centering
\includegraphics[width=1\textwidth]{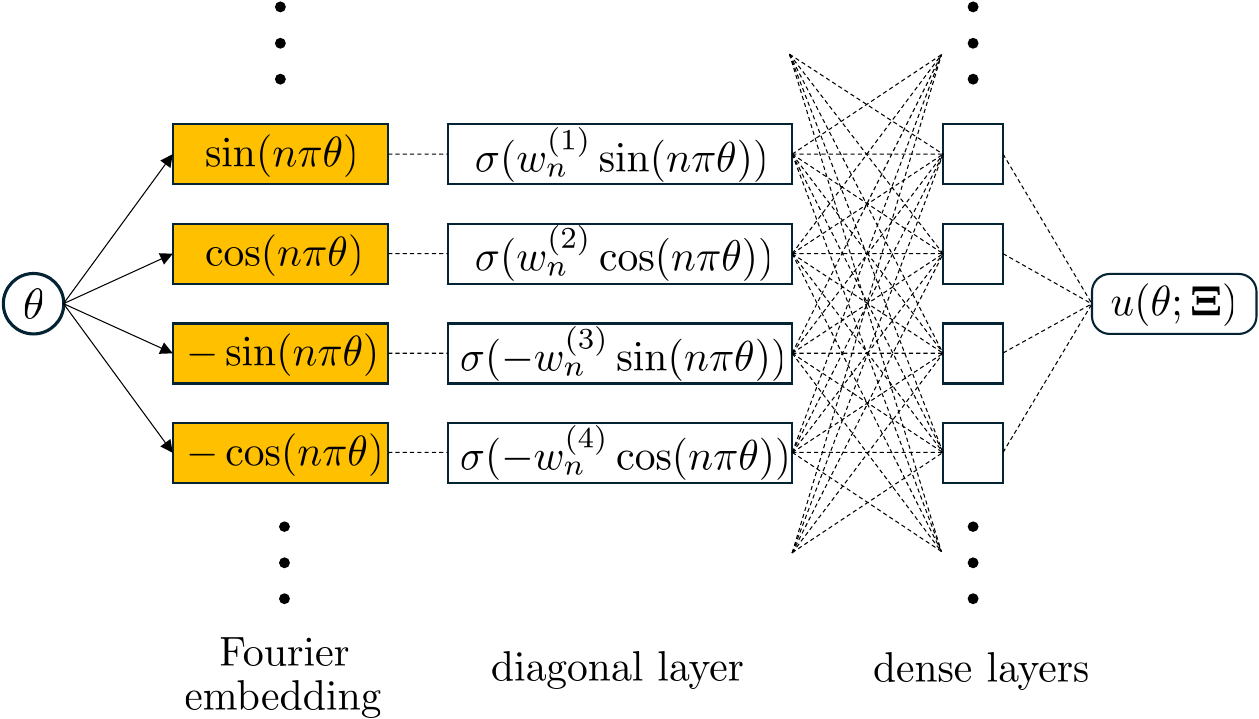}
\caption{diagonal layer subsequent to Fourier embedding}
\end{subfigure}\hspace{2mm}
\begin{subfigure}[b]{0.43\textwidth}
\centering
\includegraphics[width=1\textwidth]{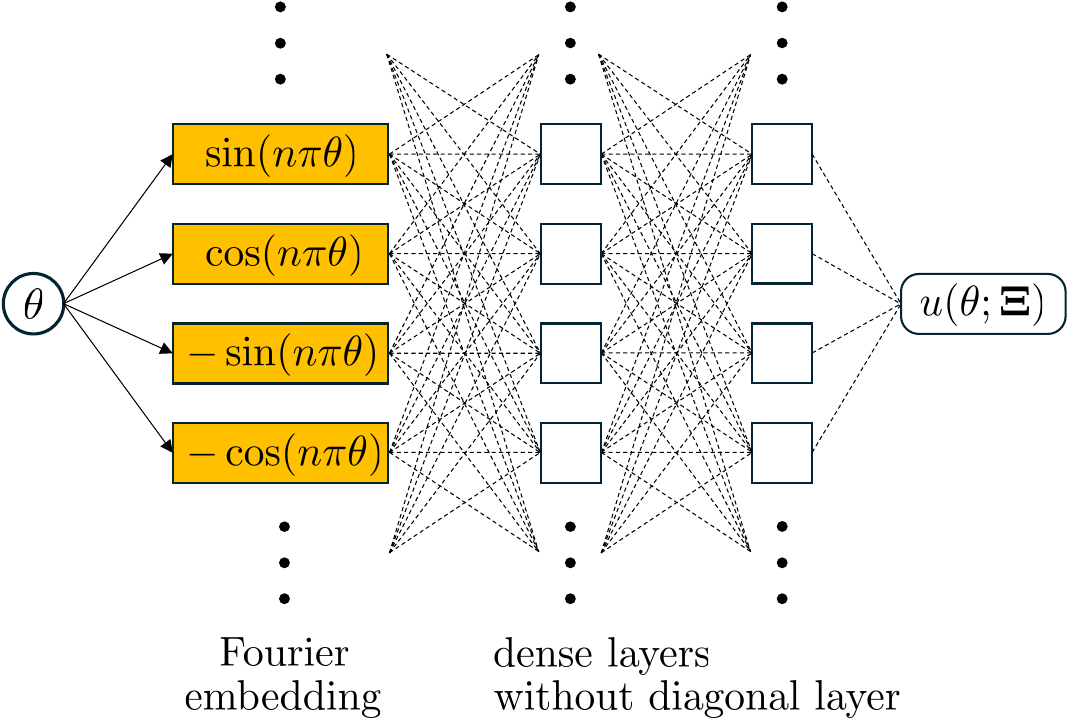}
\caption{standard Fourier embedded network}
\end{subfigure}
\caption{Fourier feature-embedded neural networks}
\label{fig:network_diagram}
\end{figure}
satisfies $\|\widetilde{c}\|_\infty \le Q_2 = O({ 8 \lambda \sqrt{m} / r^2_c })$, so $\hat{R}(w^T, \hat{c}) \le \hat{R}(w^T, \widetilde{c})$. The inequality \eqref{eq:ineq3_main_theorem} follows from applying Theorem \ref{thm:convergence_2nd_layer}. Finally, the inequality \eqref{eq:ineq4_main_theorem} follows from Lemma \ref{lem:empirical_loss_optimal_c}.
\end{proof}

By Bessel's inequality, which implies that $\|\alpha\|_\infty^2 \le \|\alpha\|_2^2 \le \|f\|_2^2 $, Theorem \ref{thm:main_theorem1} can be restated as follows. 
\begin{cor}
Under the same conditions in Theorem \ref{thm:main_theorem1}, we have
\begin{align*}
&\mathbb{E}_{\vb x, y} \left[ \ell (\hat{f}(\vb x; w^T, c^{T'}), y) \right]\\
&\le O \left({M'\|\hat{c} - c_0\|_2^2  \over T'} + {\sqrt{M' \log(1/\delta) \|\hat{c} - c_0\|_2^2 \log T'} \over \sqrt{T'} }   \right) +  {(\sum_{l=1}^s C(g_l) + \sum_{l=1}^{s'} C(h_l))^2 \over m^{2r}} \\
&\qquad + \sqrt{\nu \over TB} + 2\mathbb{E}[\epsilon^2]
 +  O \left(\left(m \|f\|_2  + \|f\|_\infty  + \kappa  \right)  \|f\|_2 \cdot  {m  \over \sqrt{TB} } \right),
\end{align*} with probability at least $1 - (T+1)\delta$.
\end{cor}

\section{Numerical Experiments}
In this section, we present a series of numerical experiments that aim to demonstrate the practical effectiveness of our proposed method in applied settings. Specifically, we aim to show how incorporating a diagonal layer after the Fourier embedding layer improves performance in regression tasks involving noisy measurements. For this, we use two distinct neural network architectures for comparison. One architecture involves the inclusion of a diagonal layer subsequent to the Fourier embedding layer as shown in \Cref{fig:network_diagram}(a), while other employs a dense layer following the embedding without the incorporation of diagonal layer as described in \Cref{fig:network_diagram}(b). For notational simplicity for further discussion, we denote these two networks $u^{\text{diag}}_{n}(\theta;\bm{\Xi})$ and $u^{\text{standard}}_n(\theta;\bm{\Xi})$, respectively, where $n$ represents the number of additional dense layers and $\bm{\Xi}$ denotes the weights of the neural network. We note that the theoretical framework established in the previous section is based on the assumption of no additional dense layers following the diagonal layer (i.e., $u^{\text{diag}}_{0}(\theta;\bm{\Xi})$). 
\begin{figure}[h!]
\centering
\includegraphics[width=0.9\textwidth]{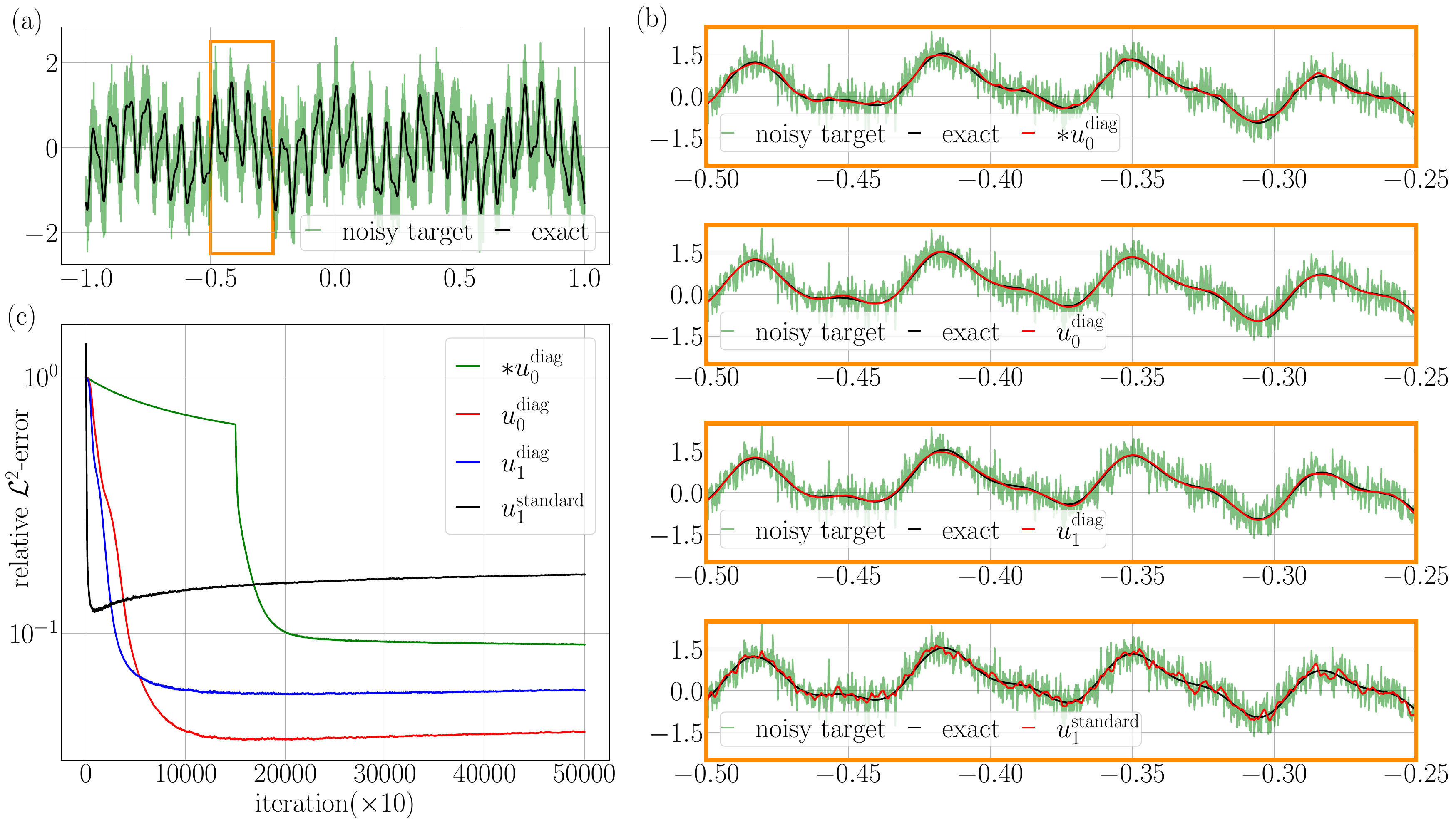}
\caption{Regression for noisy data generated by  \Cref{eq:linearFourier} using the neural networks $\ast u_0^{\text{diag}}$,$u_0^{\text{diag}}$, $u_1^{\text{diag}}$ and $u_1^{\text{standard}}$. (a): a noisy target for regression. (b): Regression results corresponding to the part in (a) enclosed by the bounding box. (c): Learning procedure during training iterations.}
\label{fig:example1_regression_result}
\end{figure}
\begin{figure}[h!]
\centering
\includegraphics[width=1.0\textwidth]{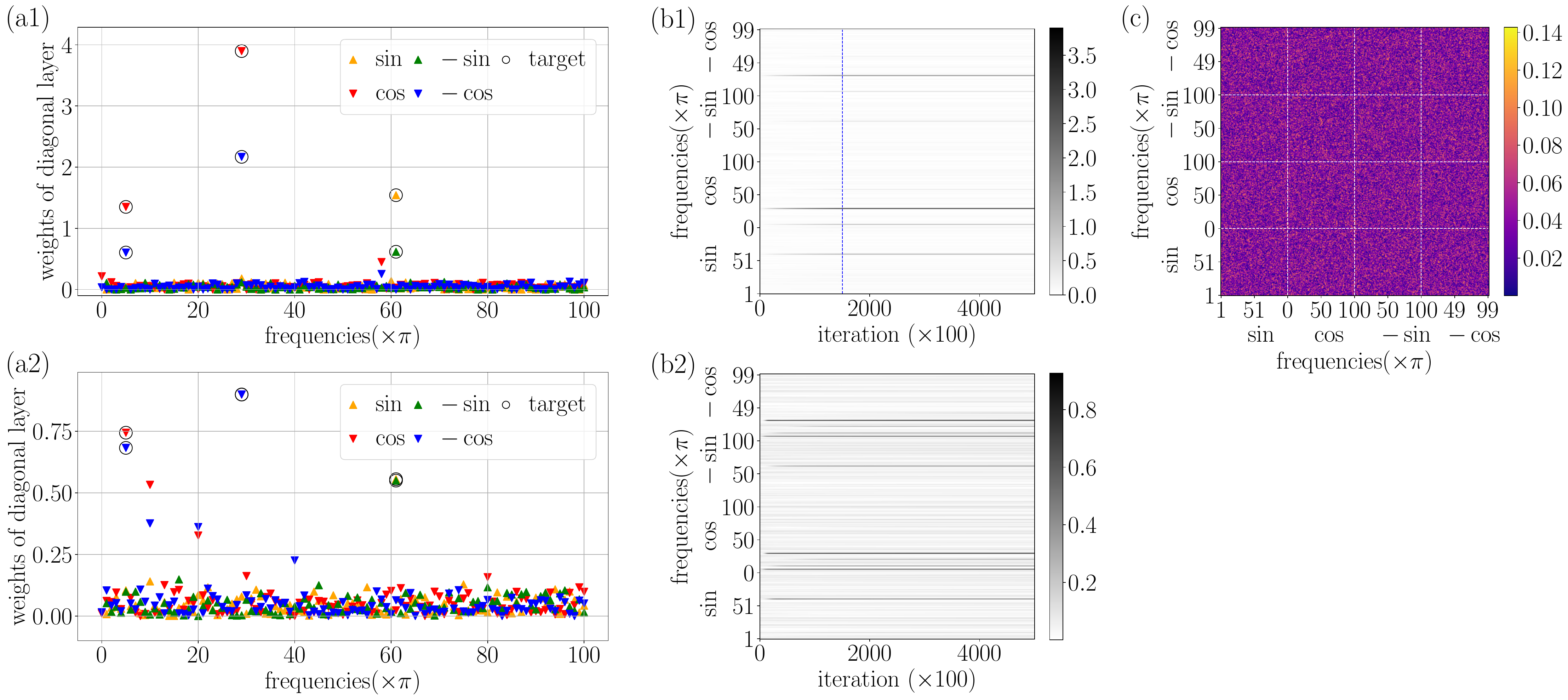}
\caption{Weight distribution of neural networks $\ast u_0^{\text{diag}}$, $u_0^{\text{diag}}$ and $u_1^{\text{standard}}$ in the regression of noisy data generated by \Cref{eq:linearFourier}. (a1) presents the final state of $\ast u_0^{\text{diag}}$ after training and (b1) is its training procedures. (a2) and (b2) correspond to $u_0^{\text{diag}}$. (c) presents the weight distribution of the dense layer subsequent to the Fourier embedding layer in  $u_1^{\text{standard}}$. }
\label{fig:example1_weights}
\end{figure}
However, our experimental findings confirm the effectiveness of this approach even when additional dense layers are included. We consider the Fourier embedding as 
\begin{equation}
\Phi(\theta)=\left[ 
\begin{matrix}
\phi(\theta) \\
-\phi(\theta)
\end{matrix}
\right]~\textrm{where}~
\phi(\theta)=\left[1,\sin(\pi \theta), \cos(\pi \theta), \cdots, \sin(m\pi \theta), \cos(m \pi \theta)\right].
\end{equation}
Here, we take into account the double signs of the Fourier modes $\pm \phi(\theta)$ 
to ensure a more comprehensive representation of the neural network equipped with only diagonal layer (i.e., $u^{\text{diag}}_{0}(\theta;\bm{\Xi})$) when employing the ReLU activation in each Fourier mode. 

The data for regression tasks comprises two types of data: 1) synthetic data generated from the deterministic function with additive Gaussian noise, which is of form $y=u^{\text{true}}(\theta)+\epsilon(\theta)$, $\theta \in \Omega$ (\Cref{subsec:linearFourier,,subsec:linearFourierPhaseShift,,subsec:nonlinearFourier}), and 2) semi-synthetic data created by incorporating synthetic noise into real-world sequential data from \textit{Google Trends} (\Cref{subsec:realWorld}).
For the cases of synthetic data, we consider the domain $\Omega=[-1,1]$ and the example data point $\theta_i$ are evenly spaced grid points on $\Omega$ with a grid size $\Delta \theta = 2\times 10^{-4}$ (i.e., $\theta_i = -1 + i \Delta \theta$, $i=0,1,2,\cdots, 10^{4}$). In the case of semi-synthetic data, the temporal domain is rescaled to $\Omega = [-1,1]$ as in the case of synthetic data.

The neural network is trained using the standard stochastic gradient descent method (SGD) with learning rate adjustments though an inverse decay function, $\alpha_m = \frac{\alpha_0}{1+\gamma m/m_0}$ where $\alpha_0$ is the initial learning rate, $\gamma$ is the decay rate, and $m_0$ is the decay steps. To initialize the network's parameters, we draw from the Glorot normal distribution \cite{glorot2010understanding} considering the differences in the number of input and output units across the layers. We conduct the training experiments using two approaches: 1) training all weights in the neural network simultaneously (i.e., standard approach) and 2) employing layer-wise training described in \Cref{alg:layerwise-training} for the network $u^{\text{diag}}_{0}(\theta;\bm{\Xi})$ specifically denote it as $\ast u^{\text{diag}}_{0}(\theta;\bm{\Xi})$. In the layer-wise training approach, the learning rate is reinitialized and decayed in a manner consistent with the training of the preceding layer. The specific values for $\alpha_0$, $\gamma$, and $m_0$ and the batch size $N_{\text{batch}}$ will be specified in each task.  

The regression performance of the approximation $u^{\text{approx}}$ is measured by the relative $\mathcal{L}^2$-error $\frac{\left\|u^{\text{approx}}-u^{\text{true}}\right\|_{2,\Omega}}{\left\|u^{\text{true}}\right\|_{2,\Omega}}$ where each $\mathcal{L}^2$-norm is calculated on an evenly spaced grid identical to the one used for training examples

\subsection{Linear function of Fourier modes}\label{subsec:linearFourier}
The first example is the synthetic data generated by a function that decomposes linearly into the three Fourier modes as follows
\begin{equation}\label{eq:linearFourier}
f(\theta) = 0.5\cos(5\pi \theta) + 0.8 \cos(29\pi \theta) + 0.3\sin(61\pi \theta) + \epsilon(\theta),~~\theta \in [-1,1].
\end{equation}
Here, the noise term $\epsilon(\theta)$ is sampled from the Gaussian distribution $\mathcal{N}(0,0.4^2)$ corresponding to the signal-to-noise ratio (SNR) $3.06$. We conduct regression using three different neural networks: $u^{\text{diag}}_{0}(\theta;\bm{\Xi})$, $u^{\text{diag}}_{1}(\theta;\bm{\Xi})$ and  $u^{\text{standard}}_{1}(\theta;\bm{\Xi})$, to assess the effectiveness of the adding a diagonal layer in regression performance. Learning procedure, with the training hyperparameters $(\alpha_0, \gamma, m_0, N_{\text{batch}})=(2\times 10^{-3}, 0.95, 5\times 10^{4}, 201)$, is presented in \Cref{fig:example1_regression_result}-(c) and the regression results after training are shown in \Cref{fig:example1_regression_result}-(b). The cases with the diagonal layer, $\ast u^{\text{diag}}_{0}(\theta;\bm{\Xi})$, $u^{\text{diag}}_{0}(\theta;\bm{\Xi})$ and $u^{\text{diag}}_{1}(\theta;\bm{\Xi})$ exhibit superior performance compared to the case without the diagonal layer $u^{\text{standard}}_{1}(\theta;\bm{\Xi})$.
These exhibited reduced sensitivity to noise, showcasing the regularization effect. Notably, despite $u^{\text{diag}}_{1}(\theta;\bm{\Xi})$ being more overparametrized than $u^{\text{standard}}_{1}(\theta;\bm{\Xi})$, 
\begin{figure}[h!]
\centering
\includegraphics[width=0.9\textwidth]{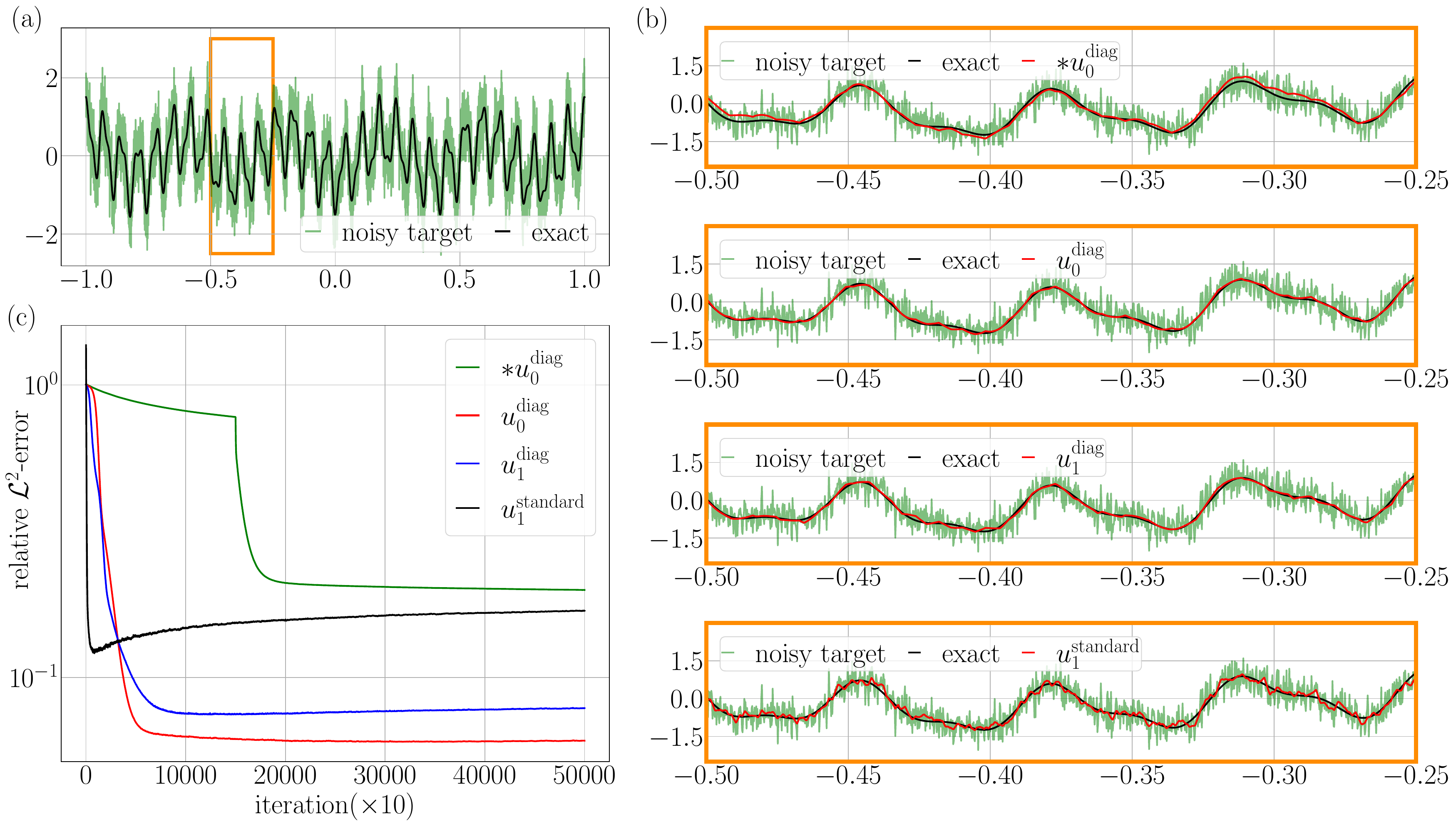}
\caption{Regression for noisy data generated by  \Cref{eq:linearFourierPhaseShift} using the neural networks $\ast u_0^{\text{diag}}$,$u_0^{\text{diag}}$, $u_1^{\text{diag}}$ and $u_1^{\text{standard}}$. (a): a noisy target for regression. (b): Regression results corresponding to the part in (a) enclosed by the bounding box. (c): Learning procedure during training iterations.}
\label{fig:example2_regression_result}
\end{figure}
\begin{figure}[h!]
\centering
\includegraphics[width=1.0\textwidth]{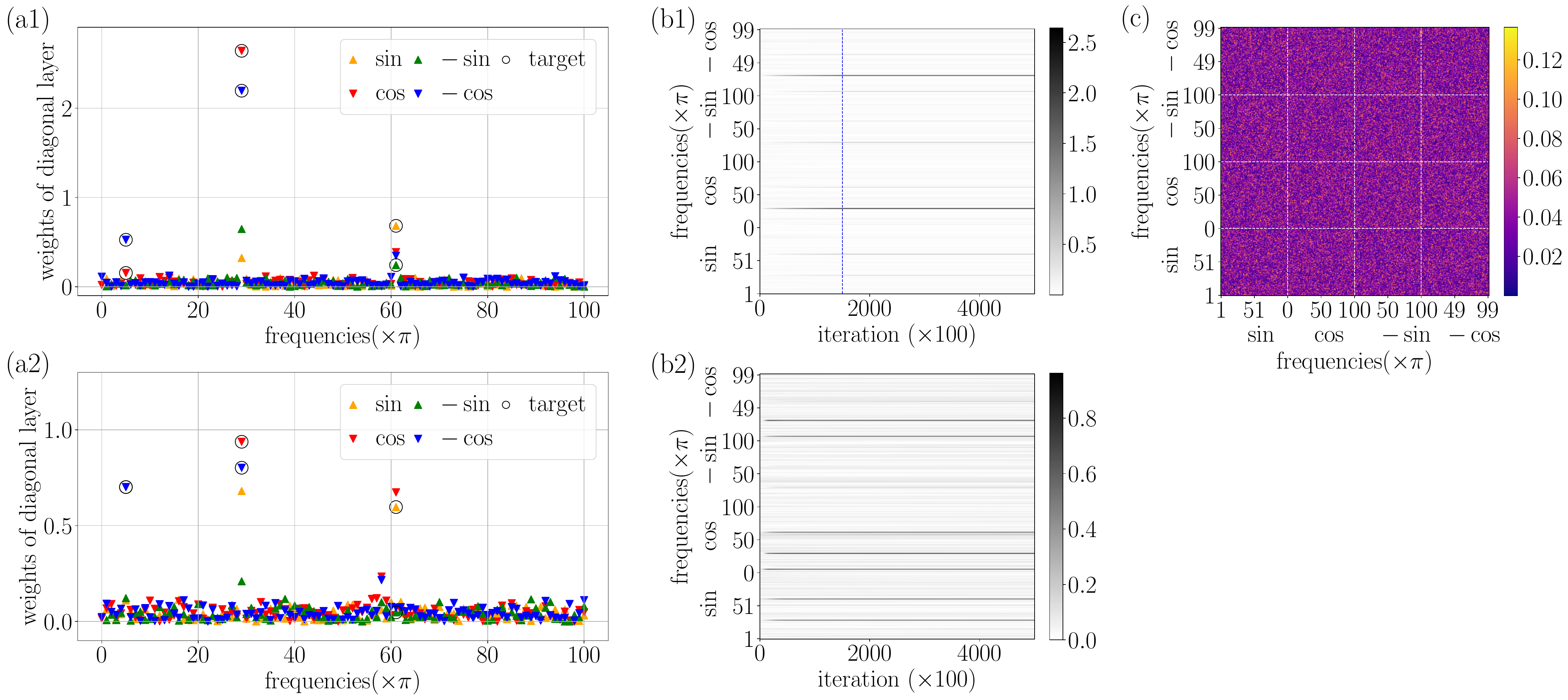}
\caption{Weight distribution of neural networks $\ast u_0^{\text{diag}}$, $u_0^{\text{diag}}$ and $u_1^{\text{standard}}$ in the regression of noisy data generated by \Cref{eq:linearFourierPhaseShift}. (a1) presents the final state of $\ast u_0^{\text{diag}}$ after training and (b1) is its training procedures. (a2) and (b2) correspond to $u_0^{\text{diag}}$. (c) presents the weight distribution of the dense layer subsequent to the Fourier embedding layer in  $u_1^{\text{standard}}$. }
\label{fig:example2_weights}
\end{figure}
it effectively mitigated overfitting to the noise. For the case without diagonal layer $u^{\text{standard}}_{1}(\theta;\bm{\Xi})$, the distribution of dense layer subsequent to the Fourier embedding layer exhibits nearly all frequencies are activated as shown in \Cref{fig:example1_weights}-(c). In contrast, the case with diagonal layers $\ast u^{\text{diag}}_{0}(\theta;\bm{\Xi})$, $u^{\text{diag}}_{0}(\theta;\bm{\Xi})$ displays activation in only a few frequencies including the target frequencies  (\Cref{fig:example1_weights}-(a1) and (a2)). This difference highlights the efficacy of the diagonal layer in preventing the overfitting of the noise. Furthermore, when comparing layer-wise training with the standard approach, it was evident that layer-wise training promoted sparsity in the distribution of the diagonal layer (\Cref{fig:example1_weights}-(a1) and (b1)) more effectively than standard training (\Cref{fig:example1_weights}-(a2) and (b2)). Interestingly, however,  simultaneous adjustments in the diagonal layer and output layer still yield superior regression performance compared to sequential learning, where the output layer is trained under fixed conditions in the diagonal layer.

\subsection{Linear function of Fourier modes with phase shift}\label{subsec:linearFourierPhaseShift}

The next example is the synthetic data as the same as the previous example except that the phase of the Fourier modes are shifted as follows
\begin{equation}\label{eq:linearFourierPhaseShift}
f(\theta) = 0.5\cos(5\pi (\theta-0.2) ) + 0.8 \cos(29\pi (\theta+0.1) ) + 0.3\sin(61\pi (\theta-0.3) ) + \epsilon(\theta).
\end{equation}
We note that the the phase-shifted modes do not precisely align across the embedding components of the neural network. As in previous example, we conduct regression using three different neural networks: $u^{\text{diag}}_{0}(\theta;\bm{\Xi})$, $u^{\text{diag}}_{1}(\theta;\bm{\Xi})$ and  $u^{\text{standard}}_{1}(\theta;\bm{\Xi})$. \Cref{fig:example1_regression_result} summarizes the regression results with training hyperparameters $(\alpha_0, \gamma, m_0, N_{\text{batch}})=(2\times 10^{-3}, 0.95, 5\times 10^{4}, 201)$. 
\noindent The effectiveness of the diagonal layer is clearly demonstrated in the case of shifted phases; it effectively regularizes and prevents overfitting to the noise. 
\begin{figure}[h!]
\centering
\includegraphics[width=0.85\textwidth]{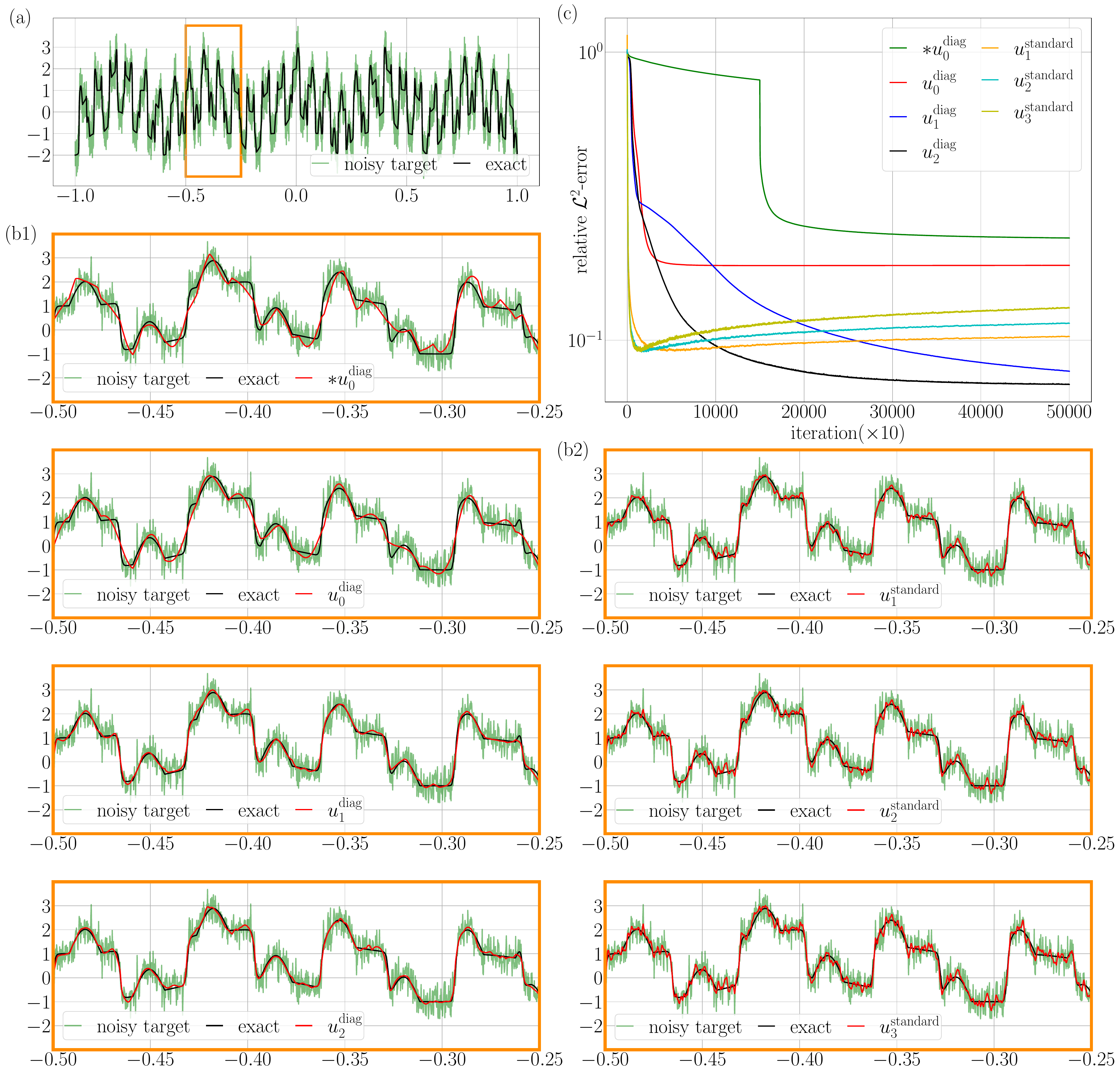}
\caption{Regression for noisy data generated by  \Cref{eq:nonlinearFourier} using the neural networks $\ast u_0^{\text{diag}}$ and $u_i^{\text{diag}}$, $i=0,1,2$ (with diagonal layer), and $u_j^{\text{standard}}$, $j=1,2,3$. (a): a noisy target for regression. (b): Regression results corresponding to the part in (a) enclosed by the bounding box. (c): Learning procedure during training iterations.}
\label{fig:example3_regression_result}
\end{figure}
Remarkably, despite the phase shifts, the distribution of the diagonal layer not only exhibits the sparsity of the Fourier modes but also captures the target frequencies (as shown in \Cref{fig:example1_weights}).  It is noteworthy that, similar to the previous example, training only the diagonal layer in a layer-wise manner can activate the target frequency modes exclusively  (\Cref{fig:example1_weights}-(b1)).

\subsection{Nonlinear function of Fourier modes}\label{subsec:nonlinearFourier}
The third example involves the synthetic data comprising Fourier modes, each nonlinearly transformed according to the equation:
\begin{equation}\label{eq:nonlinearFourier}
f(\theta) = (0.5\cos(5\pi \theta))^3 + \tanh (10 \cos(29\pi \theta)) + \max (\sin(61\pi \theta), 0)+ \epsilon(\theta).
\end{equation}
\begin{figure}[h!]
\centering
\includegraphics[width=0.95\textwidth]{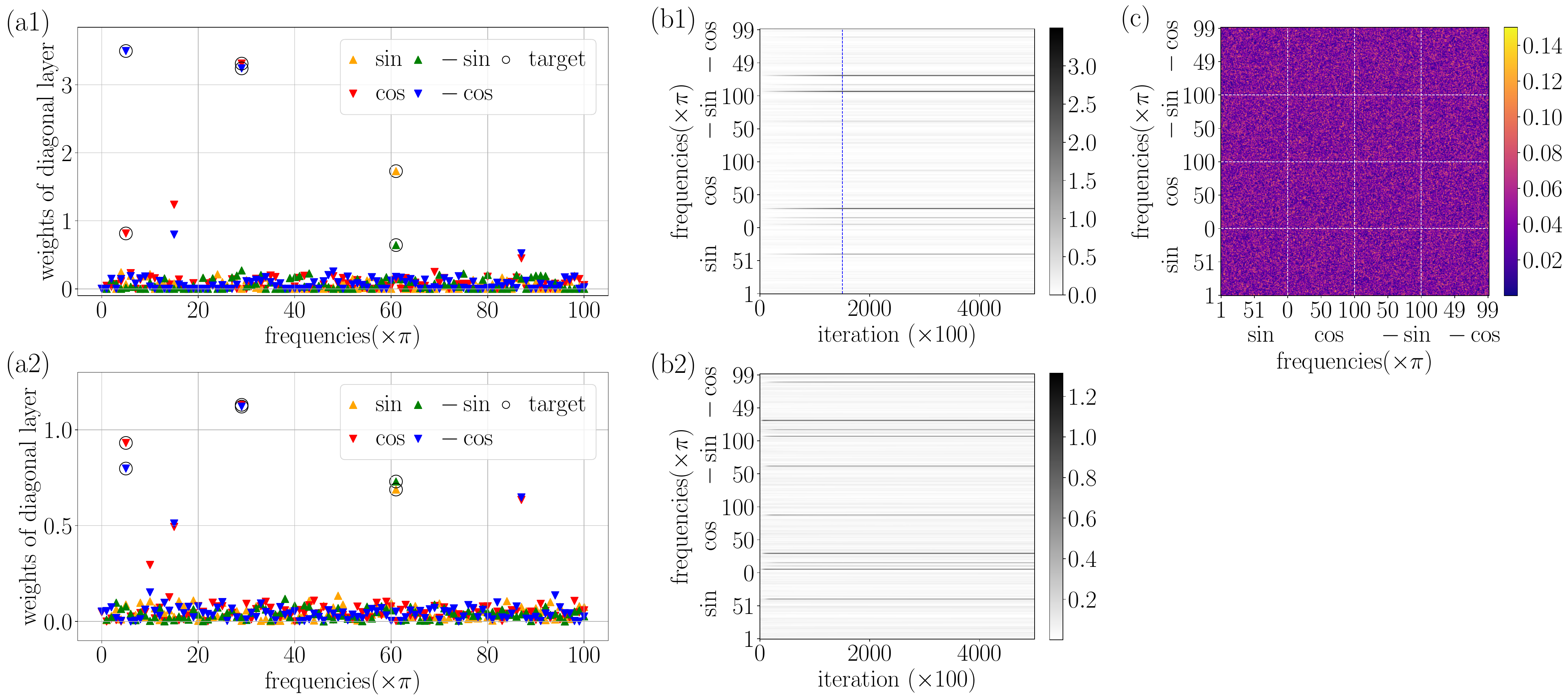}
\caption{Weight distribution of neural networks  $\ast u_0^{\text{diag}}$, $u_0^{\text{diag}}$ and $u_1^{\text{standard}}$ in the regression of noisy data generated by \Cref{eq:nonlinearFourier}. (a1) presents the final state of $\ast u_0^{\text{diag}}$ after training and (b1) is its training procedures. (a2) and (b2) correspond to $u_0^{\text{diag}}$. (c) presents weight distribution of the dense layer subsequent to the Fourier embedding layer in  $u_1^{\text{standard}}$.}
\label{fig:example3_weights}
\end{figure}

\noindent It incorporates polynomial, hyperbolic, and ReLU activation functions. Gaussian noise $\epsilon(\theta)$ is introduced, sampled from the distribution $\mathcal{N}(0,0.4^2)$ and the resulting noisy data is illustrated in \Cref{fig:example3_regression_result}-(a). The regression results of three neural networks with diagonal layers subsequent to the Fourier embedding, $u^{\text{diag}}_{i}(\theta;\bm{\Xi})$, $i=0,1,2$, and the layer-wise trained $\ast u^{\text{diag}}_{0}(\theta;\bm{\Xi})$ are showcased in \Cref{fig:example3_regression_result}-(b1). Concurrently, the standard Fourier neural networks $u^{\text{standard}}_{j}(\theta;\bm{\Xi})$, $j=1,2,3$ are presented in \Cref{fig:example3_regression_result}-(b2). 
\Cref{fig:example3_regression_result}-(b1) reveals the regularization effect of diagonal layers compared to \Cref{fig:example3_regression_result}-(b2). Notably, the network with two additional layers $u^{\text{diag}}_{2}(\theta;\bm{\Xi})$ outperform the other cases in this example as shown in \Cref{fig:example3_regression_result}-(c). These additional layers effectively represent and train the nonlinearity of the Fourier modes while balancing the regularization effect of diagonal layers. Despite the inherent nonlinearity of the data, the diagonal layer consistently captures the corresponding Fourier modes with enhanced sparsity as demonstrated in \Cref{fig:example3_weights} for the cases of $\ast u^{\text{diag}}_{0}(\theta;\bm{\Xi})$ (\Cref{fig:example3_weights}-(a1) and (b1)) and $u^{\text{diag}}_{0}(\theta;\bm{\Xi})$ (\Cref{fig:example3_weights}-(a2) and (b2)). As previous examples, all Fourier modes in the standard network $u^{\text{standard}}_{0}(\theta;\bm{\Xi})$ are activated in the dense layer subsequent to the diagonal layer as shown in \Cref{fig:example3_weights}-(c). This observation suggests that various combinations of Fourier modes can expedite training speed but also render the network susceptible to overfitting noise, as seen in \Cref{fig:example3_regression_result}-(c), where rapid convergence initially gives way to increased error due to overfitting.
\begin{figure}[h!]
\centering
\includegraphics[width=0.90\textwidth]{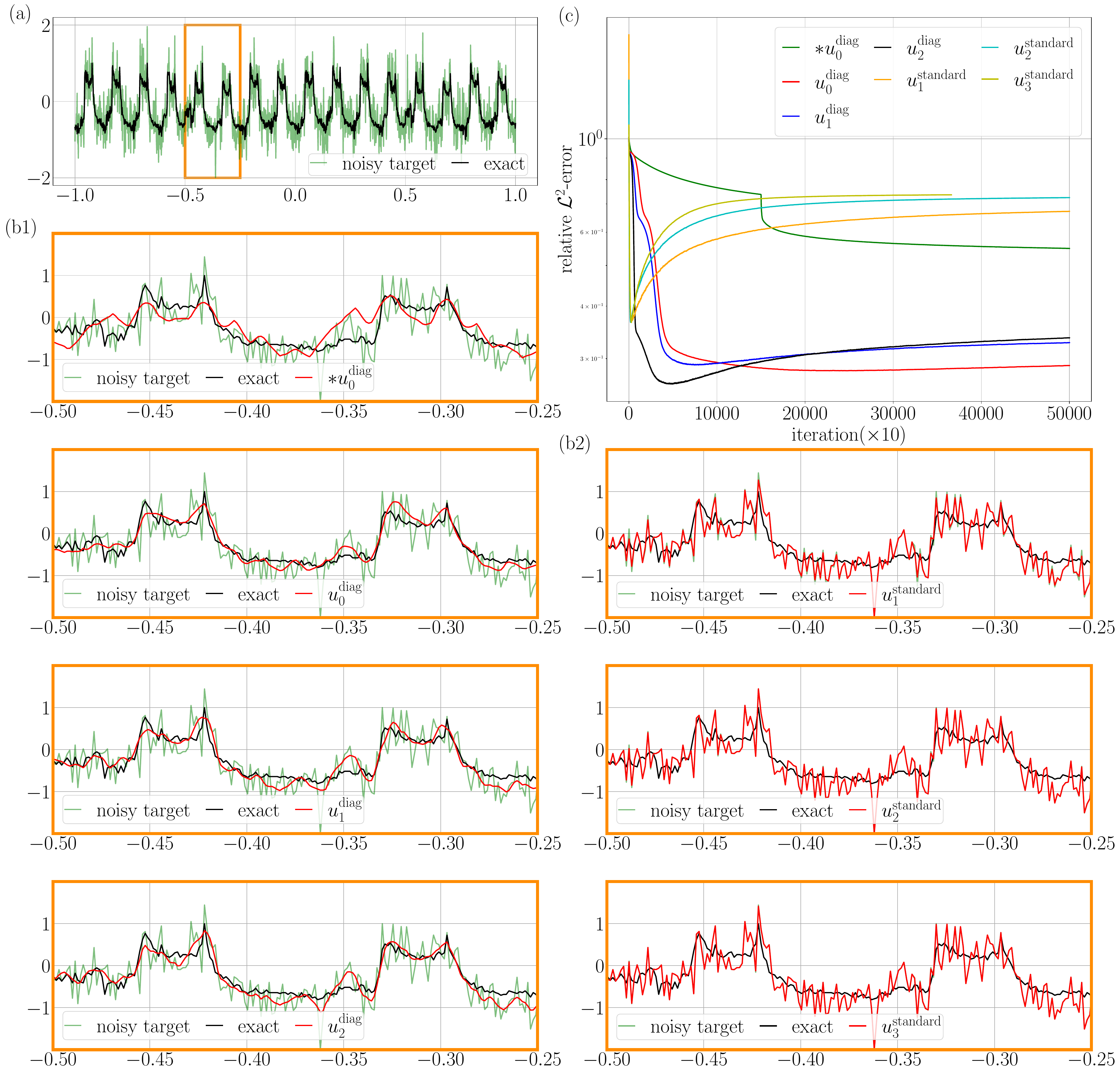}
\caption{Regression for semi-synthetic data from \textit{google trend} with the keyword `S$\&$P500' using the neural networks $\ast u_0^{\text{diag}}$ and $u_i^{\text{diag}}$, $i=0,1,2$ (with diagonal layer), and $u_j^{\text{standard}}$, $j=1,2,3$. (a): a noisy target for regression. (b): Regression results corresponding to the part in (a) enclosed by the bounding box. (c): Learning procedure during training iterations.}
\label{fig:example4_regression_result}
\end{figure}
\subsection{Real-world data}\label{subsec:realWorld}
\begin{figure}[h!]
\centering
\includegraphics[width=1.0\textwidth]{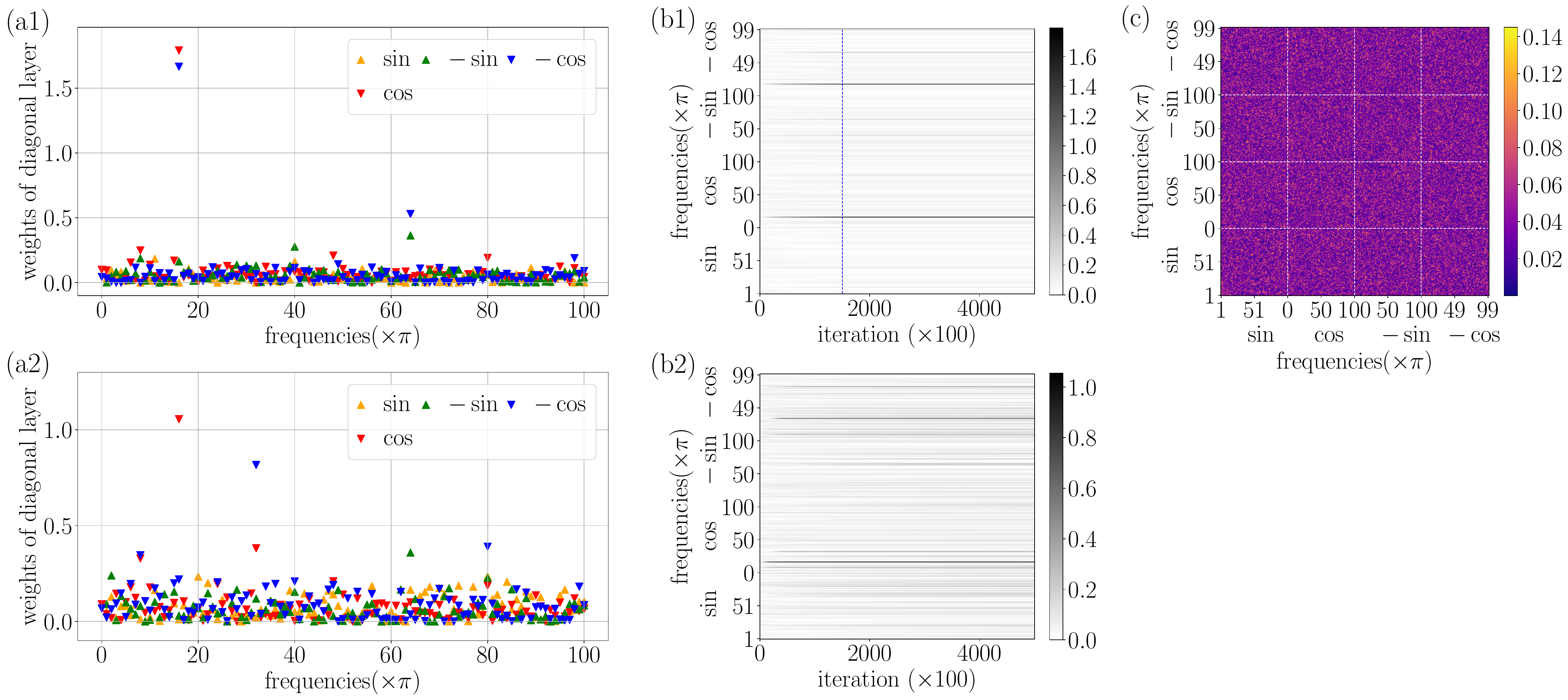}
\caption{Weight distribution of neural networks $\ast u_0^{\text{diag}}$, $u_0^{\text{diag}}$ and $u_1^{\text{standard}}$ in the regression for semi-synthetic data from \textit{google trend} with the keyword `S$\&$P500'. (a1) presents the final state of $\ast u_0^{\text{diag}}$ after training and (b1) is its training procedures. (a2) and (b2) correspond to $u_0^{\text{diag}}$. (c) presents the weight distribution of the dense layer subsequent to the Fourier embedding layer in  $u_1^{\text{standard}}$.}
\label{fig:example4_weights}
\end{figure}

The last example is the semi-synthetic data achieved from \textit{google trends} using the keyword `S$\&$P500' on 16 weekdays. We transform the temporal domain to the range $[-1,1]$ and normalize the data value within the range $[-1,1]$. Gaussian noise from $\mathcal{N}(0,0.4^2)$ is added, as depicted in \Cref{fig:example4_regression_result}-(a). The regression outcomes, employing six distinct neural networks similar to the previous example, are displayed in \Cref{fig:example4_regression_result}-(b1) (for the cases with diagonal layer) and (b2) (for the standard cases without diagonal layer). Notably, the regularization impact of the diagonal layer is evident, while the standard case exhibits overfitting to the noise. The sparse distribution of the diagonal layer enables the identification of active Fourier modes in the given dataset as illustrated in \Cref{fig:example4_weights}. Layer-wise training, in particular, results in more sparse distribution revealing that $18\pi$, $40\pi$ and $63\pi$ are active.

\section{Discussions and Conclusions} \label{sec:conclusion}
In this work, we present a novel neural network architecture that integrates Fourier embedding with diagonal layers to enhance the learning of sparse Fourier features in the presence of noise. Our approach addresses the limitations of traditional Fourier embedding, which often struggles with overfitting when labels or measurements are noisy. By introducing a simple diagonal layer after the Fourier embedding, our architecture effectively leverages the implicit regularization properties of diagonal networks, making the model more robust to noise and improving generalization.

Theoretically, we demonstrated that under certain conditions, our proposed architecture can recover the essential Fourier features of a target function and consequently learn the function with a small generalization error, even when the underlying functions are nonlinear and the signal is subject to noise. This capability is particularly significant for applications in scientific computing and machine learning, where datasets often exhibit periodic or cyclic patterns.

Our numerical experiments support the theoretical predictions, showing that the proposed method not only improves generalization performance but also inherently identifies the sparsity pattern of the target function without requiring prior knowledge of the sparsity level or extensive hyperparameter tuning. This reduces the complexity of model design and enhances the practical applicability of our approach.

\section*{Acknowledgments}
Jihun Han is supported by ONR MURI N00014-20-1-2595.

\bibliographystyle{elsarticle-num}
\bibliography{sparseFourierNN}

\clearpage
\appendix
\onecolumn
\setcounter{figure}{0} 
\renewcommand{\thefigure}{S\arabic{figure}}
\section*{Appendix: regression results for 4 examples}

\setcounter{page}{1}  
\renewcommand{\thepage}{A\arabic{page}}

\noindent
\begin{minipage}{0.44\textwidth}
    \centering
    \includegraphics[width=\textwidth]{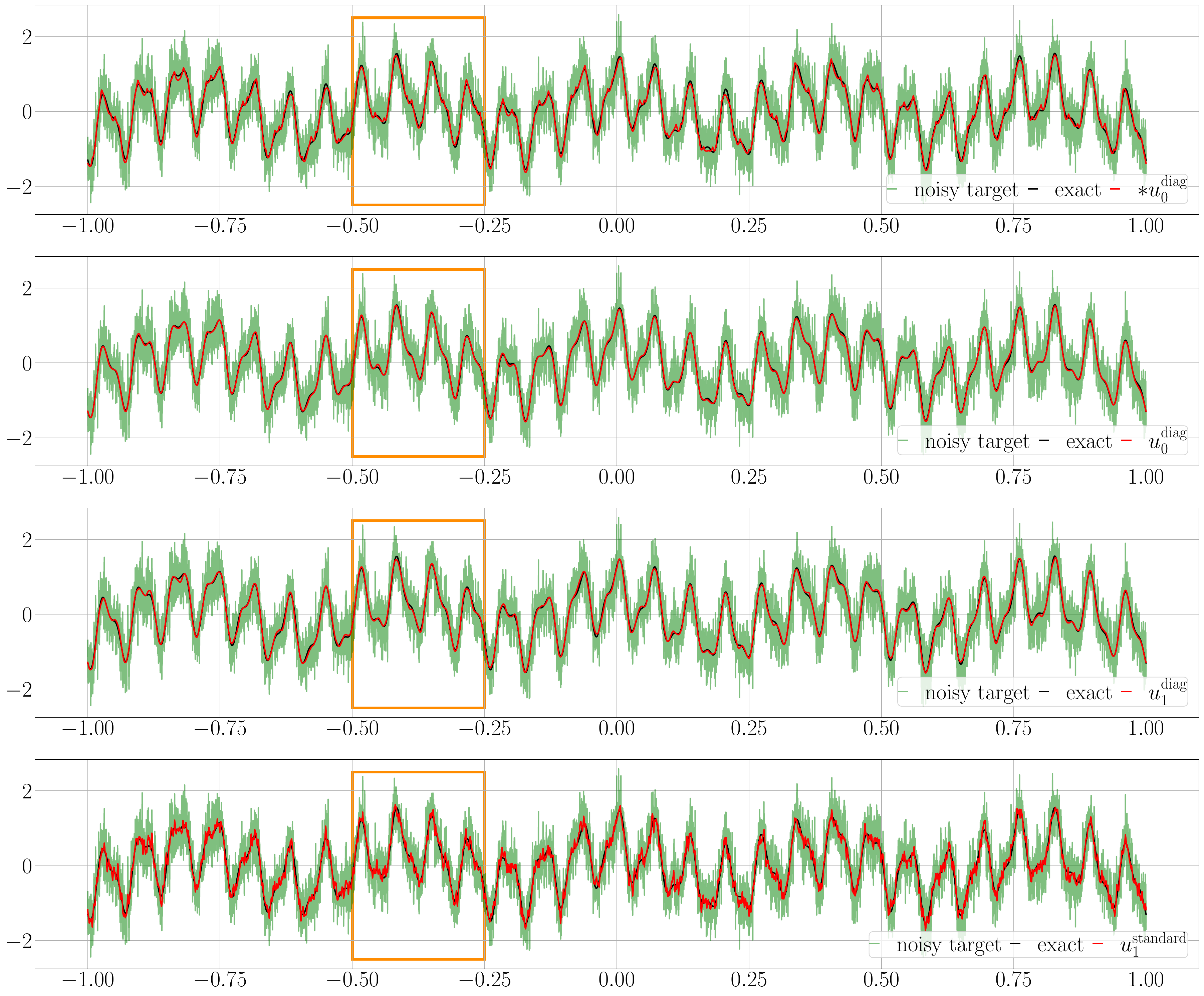}
    \captionof{figure}{Regression results for example 1 using the neural networks $\ast u_0^{\text{diag}}$,$u_0^{\text{diag}}$, $u_1^{\text{diag}}$ and $u_1^{\text{standard}}$, shown sequentially by row.}
    \label{fig:appendix_example1_regression_result}
\end{minipage} \hfill
\begin{minipage}{0.44\textwidth}
    \centering
    \includegraphics[width=\textwidth]{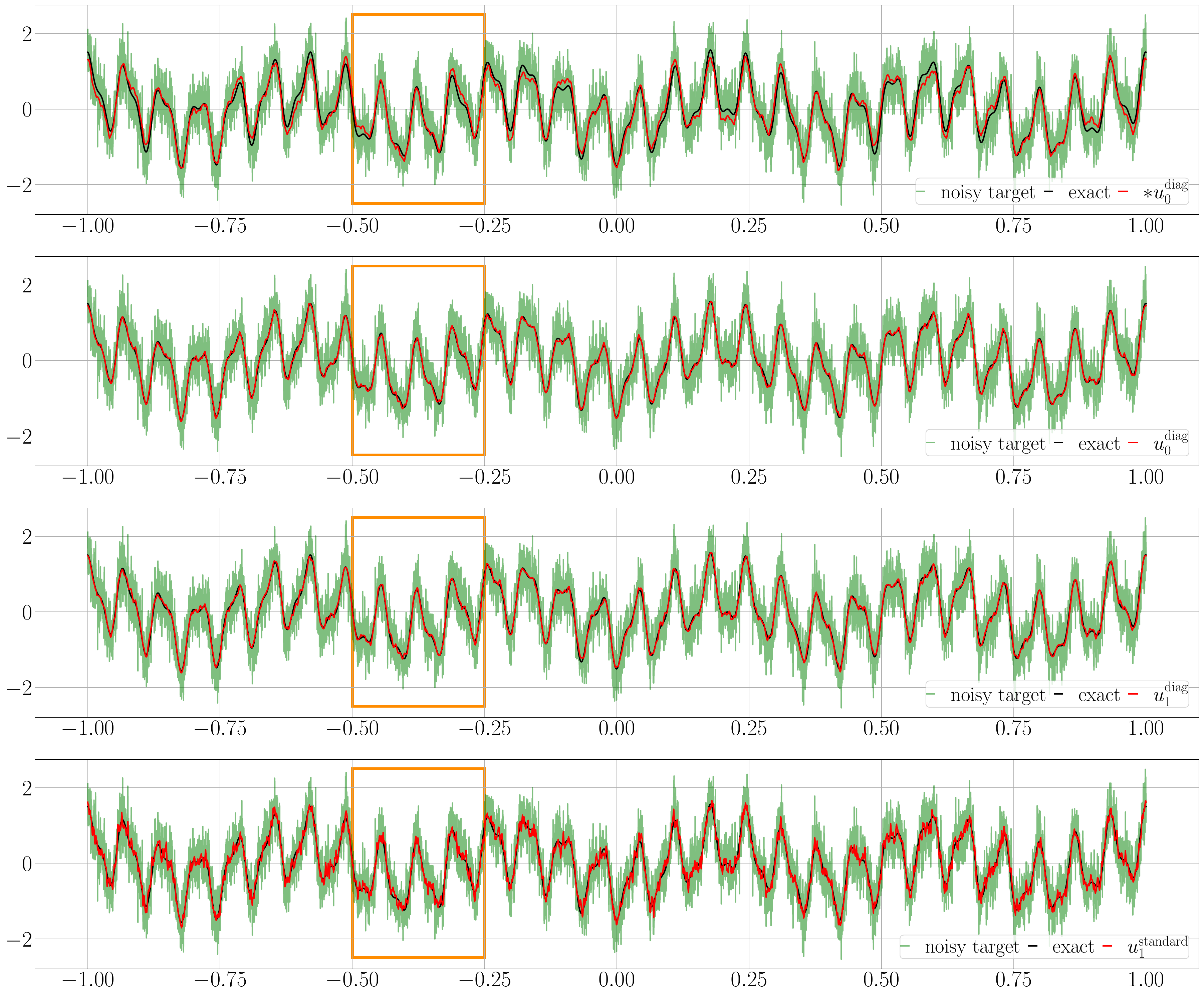}
    \captionof{figure}{Regression results for example 2 using the neural networks $\ast u_0^{\text{diag}}$,$u_0^{\text{diag}}$, $u_1^{\text{diag}}$ and $u_1^{\text{standard}}$, shown sequentially by row.}
    \label{fig:appendix_example2_regression_result}
\end{minipage}

\vspace{0.3cm}

\noindent
\begin{minipage}{0.44\textwidth}
    \centering
    \includegraphics[width=\textwidth]{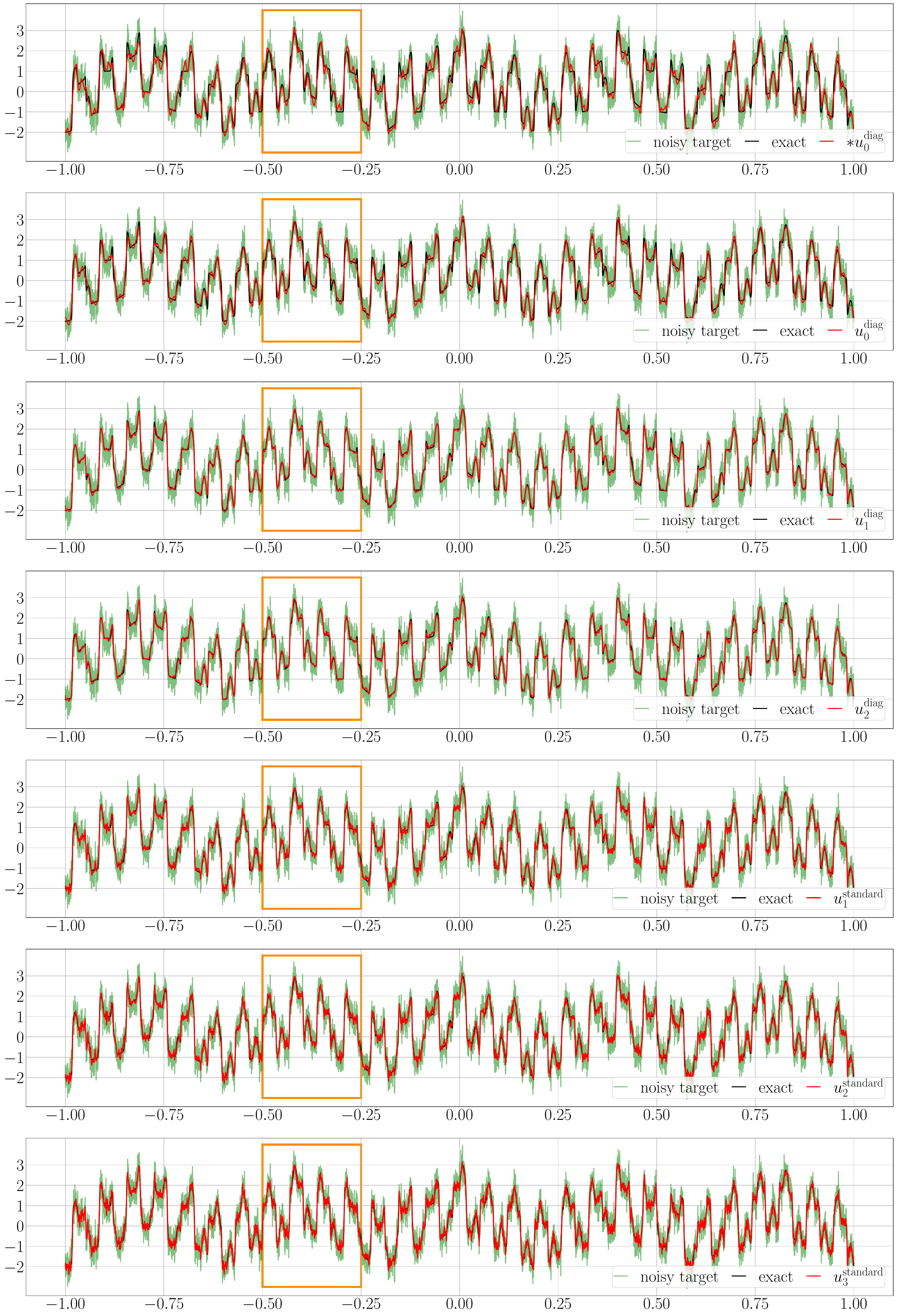}
    \captionof{figure}{Regression results for example 3 using the neural networks $\ast u_0^{\text{diag}}$ and $u_i^{\text{diag}}$, $i=0,1,2$, and $u_j^{\text{standard}}$, $j=1,2,3$, shown sequentially by row.}
    \label{fig:appendix_example3_regression_result}
\end{minipage} \hfill
\begin{minipage}{0.44\textwidth}
    \centering
    \includegraphics[width=\textwidth]{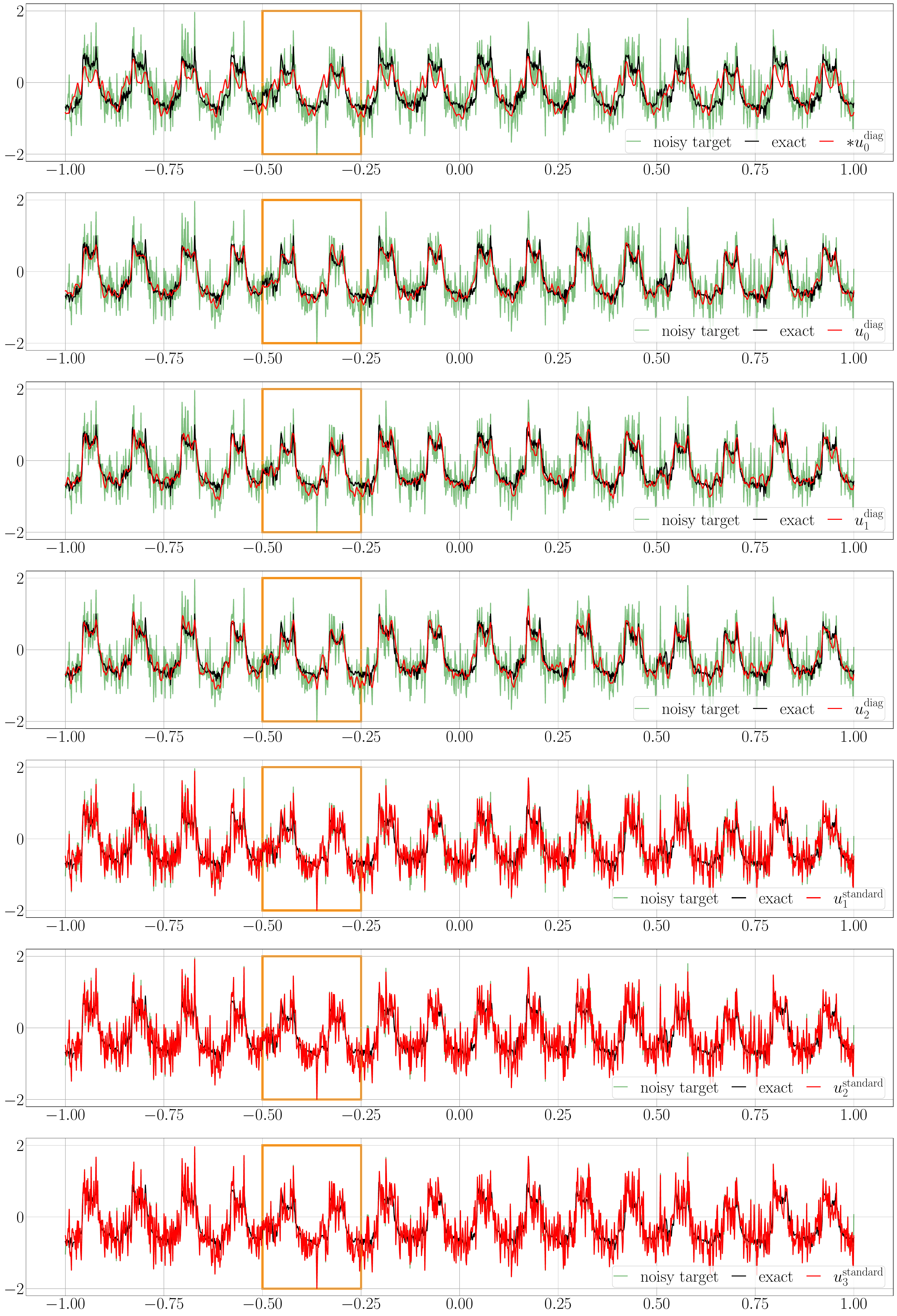}
    \captionof{figure}{Regression results for example 4 using the neural networks $\ast u_0^{\text{diag}}$ and $u_i^{\text{diag}}$, $i=0,1,2$, and $u_j^{\text{standard}}$, $j=1,2,3$, shown sequentially by row.}
    \label{fig:appendix_example4_regression_result}
\end{minipage}

\end{document}